\documentclass[10pt,journal,compsoc]{IEEEtran}
\ifCLASSOPTIONcompsoc
  \usepackage[nocompress]{cite}
\else
  \usepackage{cite}
\fi
\ifCLASSINFOpdf
\else
\fi
\usepackage{times}
\usepackage{epsfig}
\usepackage{graphicx}
\usepackage{amsmath}
\usepackage{amssymb}

\usepackage[utf8]{inputenc} 
\usepackage[T1]{fontenc}    
\usepackage{url}            
\usepackage{booktabs}       
\usepackage{amsfonts}       
\usepackage{nicefrac}       
\usepackage{microtype}      
\usepackage{algorithm}
\usepackage{array}
\usepackage{bm}
\usepackage{multirow}
\usepackage{color}

\usepackage{algpseudocode}

\def\E{{\rm E}}
\def\N{{\rm N}}

\def\tY{\tilde{Y}}

\def\hY{\hat{Y}}
\def\hX{\hat{X}}

\begin{document}
%
\title{Cooperative Training of Fast Thinking \\Initializer and Slow Thinking Solver for Conditional Learning}
%
%
%
%

\author{Jianwen Xie*, Zilong Zheng*, Xiaolin Fang, Song-Chun Zhu,~\IEEEmembership{Fellow, IEEE,}
        and~Ying Nian Wu
\IEEEcompsocitemizethanks{
\IEEEcompsocthanksitem J. Xie is with the Cognitive Computing Lab, Baidu Research, Bellevue, WA 98004, USA. E-mail: jianwen@ucla.edu
\IEEEcompsocthanksitem Z. Zheng is with the Department of Computer Science, University of California, Los Angeles, CA 90095, USA. E-mail: z.zheng@ucla.edu
\IEEEcompsocthanksitem X. Fang is with the Department of Computer Science, Massachusetts Institute of Technology, Cambridge, MA 02139, USA. E-mail: xiaolinf@csail.mit.edu
\IEEEcompsocthanksitem S.-C. Zhu is with Tsinghua University and Peking University, Beijing, China. E-mail: sczhu@stat.ucla.edu
\IEEEcompsocthanksitem Y. N. Wu is with the Department of Statistics, University of California, Los Angeles, CA 90095, USA. E-mail: ywu@stat.ucla.edu 
\IEEEcompsocthanksitem * indicates equal contributions.
}

}


\IEEEtitleabstractindextext{%
\begin{abstract}
This paper studies the problem of learning the conditional distribution of a high-dimensional output given an input, where the output and input may belong to two different domains, 
e.g., the output is a photo image and the input is a sketch image. We solve this problem by cooperative training of a fast thinking initializer and slow thinking solver. The initializer generates the output directly by a non-linear transformation of the input as well as a noise vector that accounts for latent variability in the output. The slow thinking solver learns an objective function in the form of a conditional energy function, so that the output can be generated by optimizing the objective function,  or more rigorously by sampling from the conditional energy-based model.  We propose to learn the two models jointly, where the fast thinking initializer serves to initialize the sampling of the slow thinking solver, and the solver refines the initial output by an iterative algorithm. The solver learns from the difference between the refined output and the observed output, while the initializer learns from how the solver refines its initial output. We demonstrate the effectiveness of the proposed method on various conditional learning tasks, e.g., class-to-image generation, image-to-image translation, and image recovery. The advantage of our method over GAN-based methods is that our method is equipped with a slow thinking process that refines the solution guided by a learned objective function. 
\end{abstract}

\begin{IEEEkeywords}
Deep generative models; Cooperative learning; Energy-based models; Langevin dynamics; Conditional learning.
\end{IEEEkeywords}}

\maketitle

\IEEEdisplaynontitleabstractindextext

\IEEEpeerreviewmaketitle

\IEEEraisesectionheading{\section{Introduction}\label{sec:introduction}}

\subsection{Background and motivation}
\IEEEPARstart{W}{hen} we learn to solve a problem, we can learn to directly map the problem to the solution. This amounts to fast thinking, which underlies reflexive or impulsive behavior, or muscle memory, and it can happen when one is emotional or under time constraint. We may also learn an objective function or value function that assigns values to candidate solutions,  and we optimize the objective function by an iterative algorithm to find the most valuable solution. This amounts to slow thinking, which underlies planning, search or optimal control, and it can happen when one is calm or have time to think through. 

In this paper, we study the supervised learning of the conditional distribution of a high-dimensional output given an input, where the output and input may belong to two different domains. For instance, the output may be an image, while the input may be a class label, a sketch, or an image from another domain. The input defines the {problem}, and the output is the {solution}. We also refer to the input as the {source} or {condition}, and the output as the {target}. 

We solve this problem by learning two models cooperatively. One model is an initializer. It generates the output directly by a non-linear transformation of the input as well as a noise vector, where the noise vector is to account for variability or uncertainty in the output. This amounts to fast thinking because the conditional generation is accomplished by direct mapping. The other model is a solver. It learns an objective function in the form of a conditional energy function, so that the output can be generated by optimizing the objective function, or more rigorously by sampling from the conditional energy-based model, where the sampling is  to account for variability and uncertainty. This amounts to slow thinking because the sampling is accomplished by an iterative algorithm such as Langevin dynamics \cite{neal2011mcmc}, which is an example of Markov chain Monte Carlo (MCMC) \cite{liu2008monte,barbu2020monte}. We propose to learn the two models jointly, where the initializer serves to initialize the sampling process of the solver, and the solver refines the initial solution by an iterative algorithm. The solver learns from the difference between the refined solution and the observed solution, while the initializer learns from the difference between the initial solution and the refined solution. 


\begin{figure}[h]
\centering	
\includegraphics[width=0.9\linewidth]{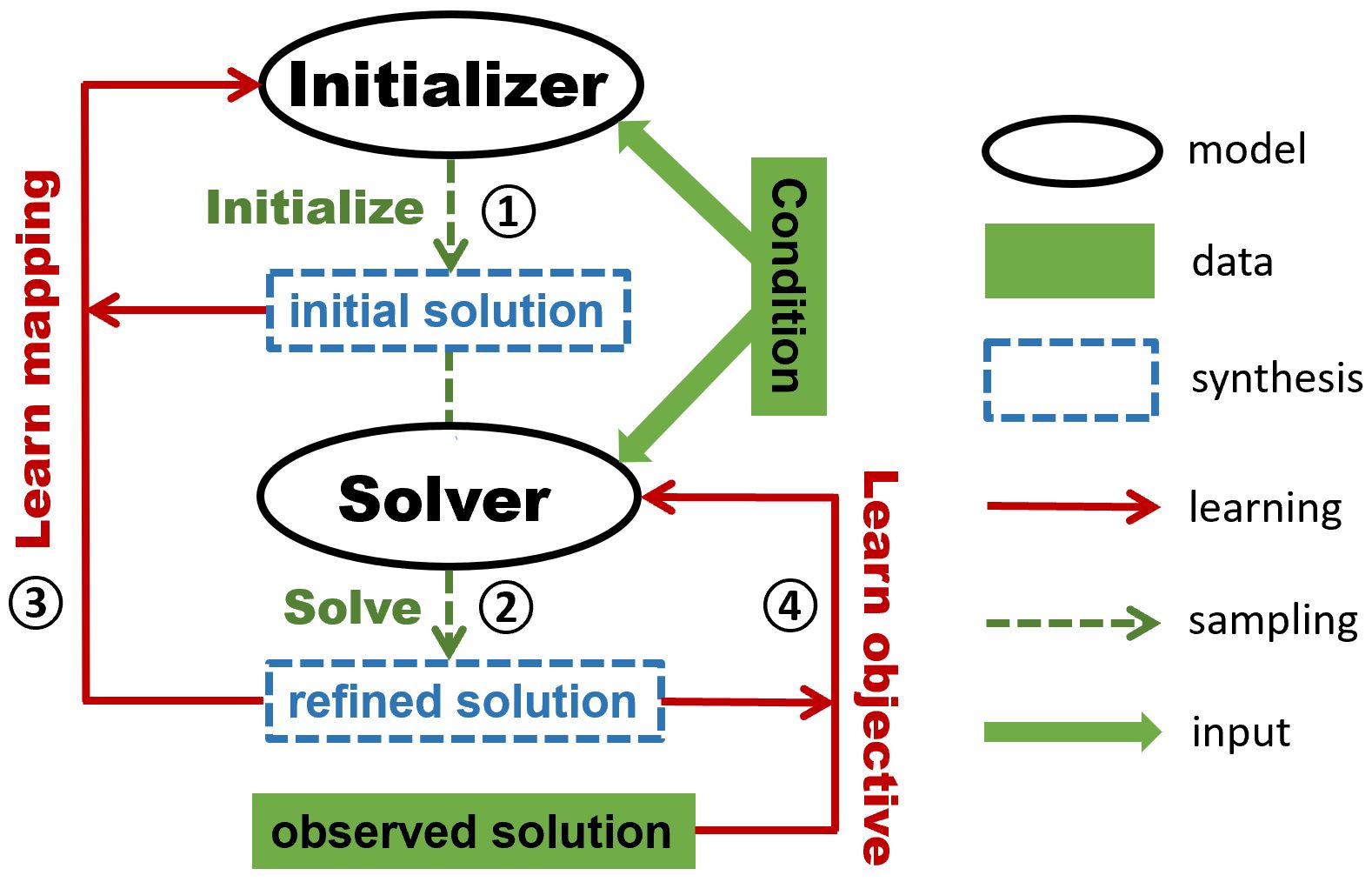} 
\caption{Diagram of fast thinking and slow thinking conditional learning. Given a condition, the initializer initializes the solver, which refines the initial solution. The initializer provides the initial solution via direct mapping (see \raisebox{.5pt}{\textcircled{\raisebox{-.9pt} {1}}}), i.e., ancestral sampling, which is a fast thinking process, while the solver refines the initial solution via Langevin sampling that optimizes the objective function (\raisebox{.5pt}{\textcircled{\raisebox{-.9pt} {2}}}), which is a slow thinking process. The initializer learns the mapping from the solver's refinement (see \raisebox{.5pt}{\textcircled{\raisebox{-.9pt} {3}}}), while the solver learns the objective function by comparing to the observed solution (see \raisebox{.5pt}{\textcircled{\raisebox{-.9pt} {4}}}). }
\label{fig:learn}
\end{figure}

Figure \ref{fig:learn} conveys the basic idea. The algorithm iterates two steps, a solving step and a learning step. The solving step consists of two stages: {\bf Initialize}: The initializer generates the initial solution according to the given condition by direct mapping, such as ancestral sampling. {\bf Solve}: The solver refines the initial solution according to the same condition by an iterative algorithm, such as Langevin sampling, which minimizes the objective function. The learning step also consists of two parts:  {\bf Learn-mapping}: The initializer updates its mapping by learning from how the solver refines its initial solution, for the purpose of providing better initial solution for the solver in the next iteration. {\bf Learn-objective}: The solver updates its objective function by shifting its high value region from the refined solution to the observed solution, for the sake of matching the refined solution to the observed one in terms of value in the next iteration. 

\begin{figure}[h]
\centering	
\includegraphics[width=0.8\linewidth]{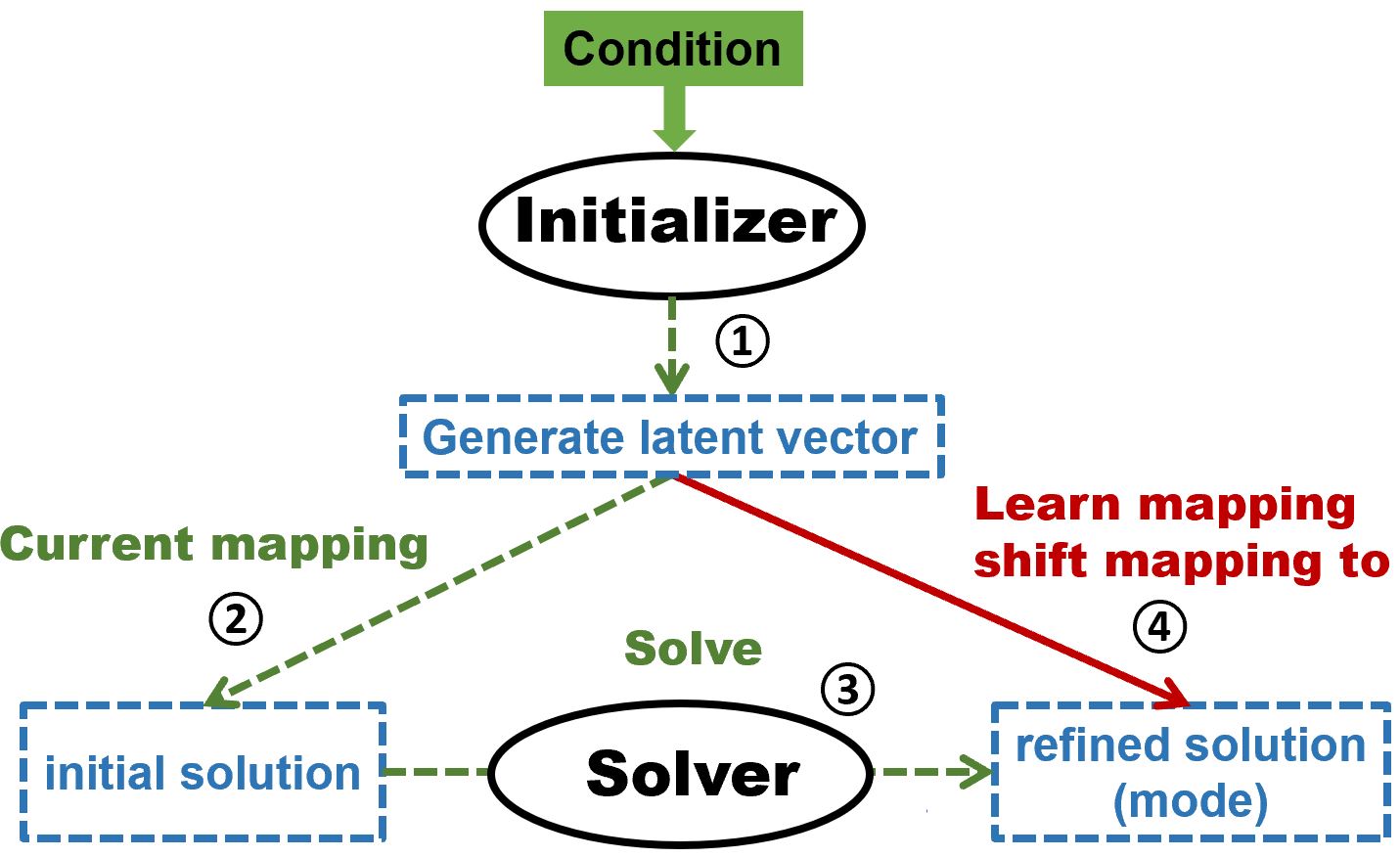} \\
(a) {\small Learn-mapping by mapping shift.  } \\ \vspace{3mm}
\includegraphics[width=0.8\linewidth]{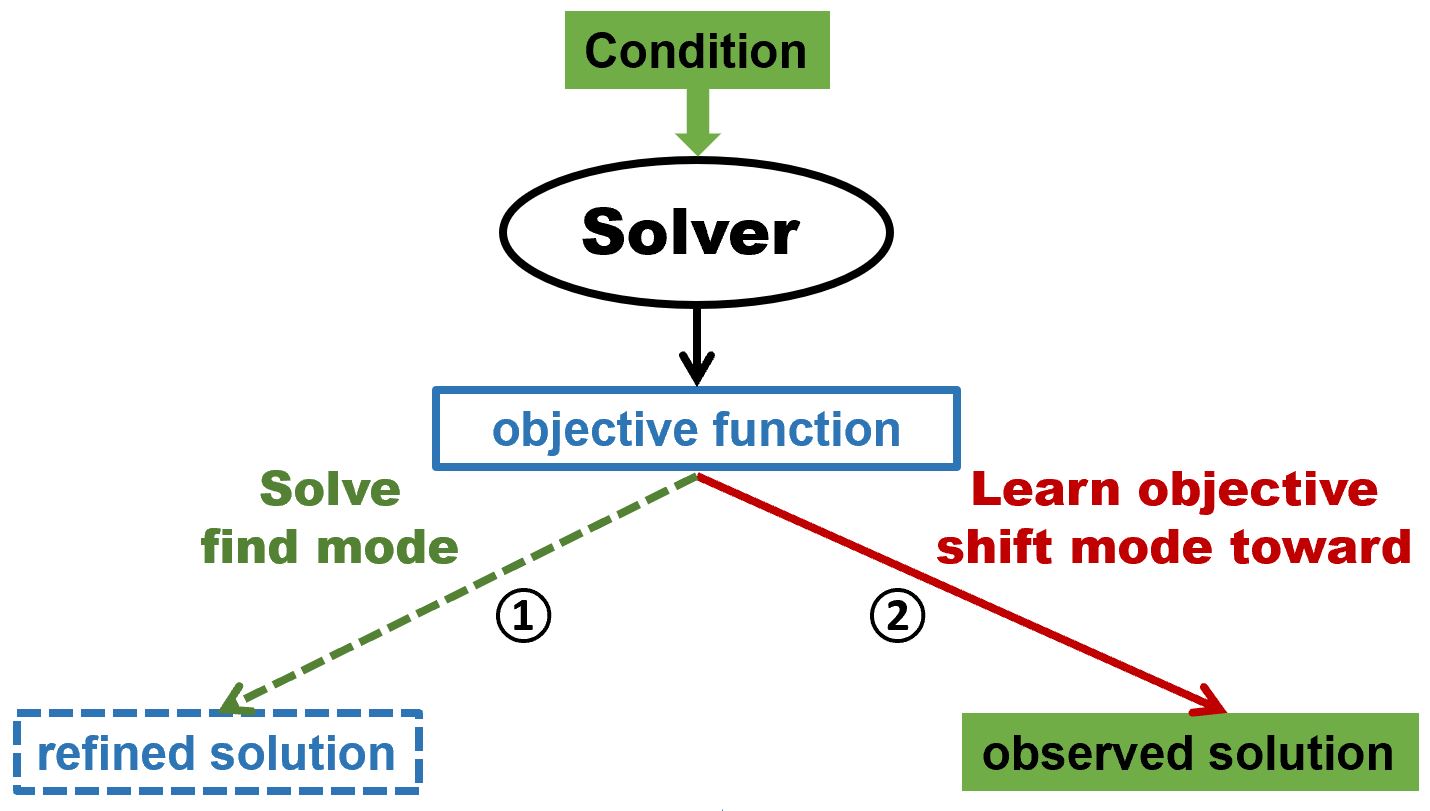} \\
(b) {\small Learn-objective by objective shift.  }
\caption{Learning step. (a) {\bf Learn-mapping by mapping shift}: In the initialize stage, the initializer generates the latent noise vector (see \raisebox{.5pt}{\textcircled{\raisebox{-.9pt} {1}}}), and maps it along with the input condition to the initial solution (see \raisebox{.5pt}{\textcircled{\raisebox{-.9pt} {2}}}). The solver outputs the refined solution after refining the initial solution (see \raisebox{.5pt}{\textcircled{\raisebox{-.9pt} {3}}}). The learning of the initializer is to shift its mapping from the initial solution toward the refined solution (see \raisebox{.5pt}{\textcircled{\raisebox{-.9pt} {4}}}). (b) {\bf Learn-objective by objective shift}: In the solve stage, the solver finds high value region or mode in its objective function via an iterative algorithm (see \raisebox{.5pt}{\textcircled{\raisebox{-.9pt} {1}}}). Those modes corresponds to the refined solution. The learning of the solver is to shift the high value region or mode of its objective function from the refined solution toward the observed solution (see \raisebox{.5pt}{\textcircled{\raisebox{-.9pt} {2}}}). }
\label{fig:learn1}
\end{figure}

Figure \ref{fig:learn1}(a) illustrates Learn-mapping step. In the Initialization step, the initializer generates the latent noise vector, which, together with the input condition, is mapped to the initial solution. In the Learn-mapping step, the initializer updates its parameters so that it maps the input condition and the latent vector to the refined solution, in order to absorb the refinement made by the solver. Because the latent vector is known, it does not need to be inferred and the learning is easy. In other words, keeping the same mapping source, the initializer shifts its mapping target from the initial solution toward the refined solution.

Figure \ref{fig:learn1}(b) illustrates Learn-objective step. In the Solve step, the solver finds the refined solution at high value region around a mode of the objective function. In the Learn-objective step, the solver updates its parameters so that the objective function shifts its high value region around the mode toward the observed solution, so that in the next iteration, the refined solution will get closer to the observed solution.

The solver shifts its mode toward the observed solution, while inducing the initializer maps the input condition and the latent vector to its mode. Learning an initializer is like mimicking ``how'', while learning a solver is like trying to understand ``why'' in terms of goal or value underlying the action. 

{\bf Why slow thinking solver?} The reason we need a solver in addition to an initializer is that it is often easier to learn the objective function than learning to generate the solution directly, since it is always easier to demand or desire something than to actually produce something directly. Because of its relative simplicity, the learned objective function can be more generalizable than the learned initializer. For instance, in an unfamiliar situation, we tend to be tentative, relying on slow thinking planning rather than fast thinking habit. 

{\bf Efficiency}. Even though we use the wording ``slow thinking'', it is only relative to ``fast thinking''. In fact, the slow thinking solver is usually fast enough, especially if it is jumpstarted by fasting thinking initializer, and there is no problem scaling up our method to big datasets. Therefore the time efficiency of the slow thinking method is not a concern. 

{\bf Student-teacher v.s. actor-critic}. We may consider the initializer as a student model, and the solver as a teacher model. The teacher refines the initial solution of the student by a refinement process, and distills the refinement process into the student. This is different from the actor-critic relationship in (inverse) reinforcement learning \cite{abbeel2004apprenticeship, ziebart2008maximum, ho2016generative} because the critic does not refine the actor's solution by a slow thinking process. 


{\bf Cooperative learning v.s. adversarial learning}. Our framework, belonging to cooperative learning \cite{xie2016cooperative,XieLuGao}, jointly learns a conditional energy-based model as the slow thinking solver and a conditional generator as the fast thinking initializer. This is essentially different from the conditional generative adversarial net (cGAN) \cite{goodfellow2014generative,isola2017image,mirza2014conditional}, where a conditional discriminator is simultaneously learned to help train the conditional generator. Our framework simultaneously trains both models and keeps both of them after training, while cGAN discards its discriminator once the generator model is well trained. In other words, our framework trains both the slow thinking solver (i.e., the energy-based model) and the fast thinking initializer (i.e., the generator), while cGAN only desires a fast thinking model (i.e., the generator). Thus, the advantage of our method over cGAN is that our method is equipped with a refinement process guided by the learned energy-based model.     

We apply our learning method to various conditional learning tasks, such as class-to-image generation, image-to-image translation, image inpainting, etc. Our experiments show that the proposed method is effective compared to other methods, such as those based on GANs \cite{goodfellow2014generative}. 

{\bf Amortized computation and temporal difference learning}. The solver is an iterative computing process. The initializer is  an amortization of this process. The learning of the initializer can be considered temporal difference learning, where the finite steps of refinements produce the temporal difference to be distilled into the initializer. 

{\bf Learning from external and internal data}. The learning of the conditional energy function is from the training data, which we may call the external data. The learning of the initializer can be considered as learning from the internal data produced by the computational process of the solver. 

{\bf Policy, value, and control}. The initializer is similar to a policy network. The solver is similar to an iterative optimal control or planning process based on a value network. The conditional energy function is similar to a cost function. 

{\bf Vector-valued initializer and scalar-valued conditional energy function}. The initializer learns a mapping from an input to a high-dimensional output. The solver learns a scalar-valued conditional energy function. It is much easier to learn a scalar-valued function than a high-dimensional vector-valued mapping, so that the iterative refinement process guided by the learned energy function improves the initializer.  

\textbf{Contributions}. This paper proposes a novel method for supervised learning of high-dimensional conditional distributions by learning a fast thinking initializer and a slow thinking solver.  We show the effectiveness of our method on conditional image generation and recovery tasks. Perhaps more importantly, 
\begin{itemize}
\item We propose a different method for conditional learning than GAN-based methods. Unlike GANs, our method has a learned value function (i.e., the energy function in the conditional energy-based model) to guide a slow thinking process to refine the solution of the initializer (i.e., conditional generator). We demonstrate the benefit of such a refinement on various image synthesis tasks. 
\item The proposed framework is generic and can be applied to a broad range of artificial intelligence problems that can be modeled via a conditional learning framework, e.g., inverse optimal control, etc. The interaction between the fast thinking initializer and the slow thinking solver can be of interest to cognitive science. 
\item This is the first paper to study conditional learning via a model-based Initializer-solver framework. It is fundamental and important to AI community.
\end{itemize}

\subsection{Related work}
The following themes are closely related to our research. We will briefly review each of them and connect them with our work.  
 
\textit{Conditional adversarial learning}. Generative Adversarial Networks (GANs) \cite{goodfellow2014generative} proposed by Goodfellow et al. have demonstrated promising results of image generation in \cite{radford2015unsupervised}, which belongs to unconditional learning, in which no supervision signals are used. With the success of adversarial learning, the conditional version of GAN (i.e., conditional GAN or cGAN) \cite{reed2016generative} has become a popular framework for supervised conditional learning, and it has been successfuly appplied to different scenarios that can be modeled in the context of conditional learning. For example, \cite{mirza2014conditional,denton2015deep} use conditional GANs for image synthesis based on class labels.  \cite{reed2016generative,zhang2017stackgan} study text-conditioned image synthesis. Other examples include image-to-image translation \cite{isola2017image}, semantic-image-to-photo translation \cite{wang2018high}, super-resolution \cite{ledig2017photo}, and video-to-video synthesis \cite{wang2018video}, etc. Our work studies similar problems. The major difference between the conditional GAN and our method is that ours is based on a conditional energy function that serves as an objective function and an iterative algorithm, which is the Langevin dynamics  guided by this objective function. This iterative process corresponds to slow thinking. Existing adversarial learning methods
do not involve this slow thinking refinement process. 

\textit{Cooperative learning}. Just as the conditional GAN is inspired by the original GAN \cite{goodfellow2014generative}, our learning method is inspired by the recent work of generative cooperative networks (CoopNets) \cite{xie2016cooperative,XieLuGao}, where the models are unconditioned. Specifically, the CoopNets framework consists of an unconditional energy-based model and an unconditional latent variable model, and jointly trains both models via MCMC teaching \cite{xie2016cooperative}, where the latent variable model learns to initialize the MCMC sampling of the energy-based model. While unconditioned generation is interesting, conditional generation and recovery is much more useful in applications. It is also much more challenging because we need to incorporate the input condition into both the initializer and the solver. Thus our method is a substantial generalization of the CoopNets\cite{xie2016cooperative}, and our extensive experiments convincingly demonstrate the usefulness of our method, which in the meantime provides a different methodology from GAN-based methods. Our work is the first to study conditional cooperative learning, and propose the fast thinking and slow thinking framework as a conditional version of CoopNets.

\textit{Conditional random field}. The objective function and the conditional energy-based model can also be considered a form of conditional random field \cite{lafferty2001conditional}. Unlike traditional conditional random field, our conditional energy function is defined by a trainable deep network, and its MCMC sampling process is jumpstarted by a non-iterative initializer. 


\textit{Energy-based generative neural nets}. Our slow thinking solver is related to energy-based generative neural nets \cite{XieLuICML,gao2018learning,XieCVPR17,xie2019learning,xie2018learning,nijkamp2019learning,nijkamp2019anatomy}, which are energy-based models (EBMs) with energy functions parameterized by deep neural nets, and trained by MCMC-based maximum likelihood learning. \cite{XieLuICML} is the first to learn EBMs parametrized by modern ConvNets by maximum likelihood estimation via Langevin dynamics, and also investigates ReLU \cite{krizhevsky2012imagenet} with Gaussian reference in the proposed model that is called generative ConvNet. \cite{gao2018learning} proposes a multi-grid sampling and learning method for training generative ConvNets. The spatial-temporal generative ConvNet proposed in\cite{XieCVPR17,xie2019learning} further generalizes the generative ConvNet of images in \cite{XieLuICML} to modeling dynamic patterns, e.g., videos or image sequences, by parameterizing the energy function with a bottom-up spatial-temporal ConvNet. \cite{xie2018learning, xie2020generative} develops a volumetric version of the energy-based generative neural net, which is called generative VoxelNet, for 3D object patterns. Recently, \cite{nijkamp2019learning} investigates training the energy-based generative ConvNet with a short-run MCMC. All models mentioned above are unconditioned EBMs, while our solver is a conditioned EBM jointly trained with a conditional latent variable model serving as an approximate conditional sampler. 

\textit{Inverse reinforcement learning}. Our method is related to inverse reinforcement learning and inverse optimal control \cite{abbeel2004apprenticeship,ziebart2008maximum}, where the initializer corresponds to the policy, and the solver corresponds to the planning or optimal control.  Unlike the action space in reinforcement learning, the output in our work is of a much higher dimension, a fact that also distinguishes our work from common supervised learning problem such as classification. As a result, the initializer needs to transform a latent noise vector (along with an input condition) to generate the initial solution, and this is different from the policy in reinforcement learning, where the policy is defined by the conditional distribution of action given state, without resorting to a latent vector. 

\textit{Unsupervised conditional learning}. Some methods study unsupervised conditional learning, where the inputs and outputs are unpaired in the training set. For example, CycleGAN \cite{zhu2017unpaired} jointly trains two GANs and enforces a cycle-consistency regularization between them to learn a two-way translator between two image collections in the absence of paired examples. AlignFlow \cite{grover2020alignflow} adopts normalizing flow models \cite{dinh2014nice,dinh2016density} to solve this problem. Recently, CycleCoopNets \cite{xie2021cycle} tackles the unpaired translation problem based on the framework of cooperative learning. Our work belongs to supervised conditional learning, where the correspondence between source domain and target domain is given and used as supervision during training.  

\section{Cooperative conditional learning}
  
Let $Y$ be the $D$-dimensional output signal of the target domain, and $C$ be the input signal of the source domain, where ``C'' stands for ``condition''. $C$ defines the problem, and $Y$ is the solution. Our goal is to learn the conditional distribution $p(Y|C)$ of the target signal (solution) $Y$ given the source signal $C$ (problem) as the condition. We shall learn $p(Y|C)$ from the training dataset of the pairs $\{(Y_i, C_i), i = 1, ..., n\}$ with the fast thinking initializer and slow thinking solver.  

\subsection{Slow thinking solver} 

The solver is based an objective function or value function $f(Y, C; \theta)$ defined on $(Y, C)$. $f(Y, C; \theta)$ can be parametrized by a bottom-up convolutional network (ConvNet) where $\theta$ collects all the weight and bias parameters. Serving as a negative energy function, $f(Y, C; \theta)$ defines a joint energy-based model \cite{XieLuICML}: 
\begin{eqnarray} 
   p(Y,C; \theta) = \frac{1}{Z(\theta)} \exp\left[ f(Y,C; \theta)\right], 
\end{eqnarray}
where $Z(\theta) = \int  \exp\left[ f(Y,C; \theta)\right] dY dC$ is the normalizing constant. 

Fixing the source signal $C$, $f(Y, C; \theta)$ defines the value of the solution $Y$ for the problem defined by $C$, and $-f(Y, C; \theta)$ defines the conditional energy function. The conditional probability is given by 
\begin{eqnarray} 
   p(Y|C; \theta) = \frac{p(Y,C;\theta)}{p(C;\theta)}
   =\frac{p(Y,C;\theta)} {\int p(Y,C; \theta) dY} \nonumber\\ 
   =\frac{1}{Z(C, \theta)} \exp \left[ f(Y,C; \theta)\right], \label{D}
\end{eqnarray}  
where $Z(C, \theta)=Z(\theta)p(C;\theta)$.  The learning of this model seeks to maximize the conditional log-likelihood function 
\begin{eqnarray}
   L(\theta) = \frac{1}{n} \sum_{i=1}^{n} \log p(Y_i|C_i; \theta), 
\end{eqnarray}
whose gradient $L'(\theta)$ is 
\begin{eqnarray}
\sum_{i=1}^n \left\{ \frac{\partial}{\partial \theta} f(Y_i, C_i; \theta) - \E_{p(Y|C_i, \theta)} \left[\frac{\partial}{\partial \theta} f(Y, C_i; \theta)\right]\right\}, \label{eq:g_L}
\end{eqnarray}  
where $\E_{p(Y|C; \theta)}$ denotes the expectation with respect to $p(Y|C, \theta)$. 
The identity underlying (\ref{eq:g_L}) is 
$\frac{\partial}{\partial \theta} \log Z(C, \theta) = \E_{p(Y|C, \theta)}\left[\frac{\partial}{\partial \theta} f(Y, C; \theta) \right]. $

The expectation in  (\ref{eq:g_L}) is analytically intractable and can be approximated by drawing samples from $p(Y|C, \theta)$ and then computing the Monte Carlo average. This can be solved by an iterative algorithm, which is a slow thinking process. One solver is the Langevin dynamics for sampling $Y \sim p(Y|C, \theta)$. It iterates the following step: 
\begin{eqnarray}
   Y_{\tau+1} = Y_\tau + \frac{\delta^2}{2}  \frac{\partial}{\partial Y} f(Y_\tau, C; \theta)  + \delta U_{\tau},  \label{eq:LangevinD}
\end{eqnarray}
where $\tau$ indexes the time steps of the Langevin dynamics, $\delta$ is the step size, and $U_{\tau} \sim \N(0, I_D)$ is Gaussian white noise. $D$ is the dimensionality of $Y$. A Metropolis-Hastings acceptance-rejection step can be added to correct for finite $\delta$. The Langevin dynamics is gradient descent on the energy function, plus noise for diffusion so that it samples the distribution instead of being trapped in the local modes. 

For each observed condition $C_i$, we run the Langevin dynamics according to (\ref{eq:LangevinD}) to obtain the corresponding synthesized example $\tY_i$ as a sample from $p(Y|C_i, \theta)$. The Monte Carlo approximation to $L'(\theta)$ is 
\begin{eqnarray} 
  L'(\theta) \approx   \frac{\partial}{\partial \theta} \left[ \frac{1}{n} \sum_{i=1}^{n} f(Y_i, C_i; \theta) - \frac{1}{n} \sum_{i=1}^{n} f(\tY_i, C_i; \theta)\right]. \label{eq:learningD}
\end{eqnarray} 
 We can then update $\theta^{(t+1)}=\theta^{(t)}+\gamma_t L'(\theta^{(t)})$. 
 
 {\bf Objective shift}: The above gradient ascent algorithm is to increase the average value of the observed solutions versus that of the refined solutions, i.e., on average, it shifts high value region or mode of $f(Y, C_i; \theta)$ from the generated solution $\tY_i$ toward the observed solution $Y_i$.  
 
 The convergence of such a stochastic gradient ascent algorithm has been studied by \cite{younes1999convergence}.

\subsection{Fast thinking initializer}

The initializer is of the following form: 
\begin{eqnarray} 
  X \sim \N(0, I_d),  \;  Y = g(X, C; \alpha) + \epsilon,  \; \epsilon \sim \N(0, \sigma^2  I_D),   \label{eq:G}
 \end{eqnarray}
where $X$ is the $d$-dimensional latent noise vector,  and $g(X,C; \alpha)$ is a top-down ConvNet defined by the parameters $\alpha$. The ConvNet $g$  maps the observed condition $C$ and the latent noise vector $X$ to the signal $Y$ directly. 
If the source signal $C$ is of high dimensionality, we can parametrize $g$ by an encoder-decoder structure: we first encode $C$ into a latent vector $Z$, and then we map $(X, Z)$ to $Y$ by a decoder. 
Given $C$, we can generate $Y$ from the conditional generator model by direct sampling, i.e., first sampling $X$ from its prior distribution, and then mapping $(X, Z)$ into $Y$ directly. This is fast thinking without iteration. 

We can learn the initializer from the training pairs $\{(Y_i, C_i), i = 1, ..., n\}$ by maximizing the conditional log-likelihood $L(\alpha) = \frac{1}{n} \sum_{i=1}^{n} \log p(Y_i|C_i, \alpha)$, where $p(Y|C, \alpha) = \int p(X) p(Y|C, X, \alpha) dX$. The learning algorithm iterates the following two steps. (1) Sample $X_i$ from $p(X_i|Y_i, C_i, \alpha)$ by Langevin dynamics. (2) Update $\alpha$ by gradient descent on $\frac{1}{n} \sum_{i=1}^{n} \|Y_i - g(X_i, C_i; \alpha)\|^2$. See \cite{HanLu2016} for details. 

\subsection{Cooperative training of initializer and solver}

The initializer and the solver can be trained jointly as follows. 

(1) The initializer supplies initial samples for the MCMC of the  solver.  For each observed condition input $C_i$, we first generate $\hX_i \sim \N(0, I_d)$, and then generate the initial solution $\hY_i=g(\hX_i, C_i; \alpha)+\epsilon_i$. If the current initializer is close to the current solver, then the generated $\{\hY_{i}, i=1,...,n\}$ should be a good initialization for the solver to sample from $p(Y|C_i, \theta)$, i.e., starting from the initial solutions $\{\hY_i, i=1,...,n\}$, we run Langevin dynamics for $l$ steps to get the refined solutions $\{\tilde{Y_i}, i=1,...,n\}$. These $\{\tilde{Y_i}\}$ serve as the synthesized examples from $p(Y|C_i)$ and are used to update $\theta$ in the same way as we learn the solver model in equation (\ref{eq:learningD}) for objective shifting. 

(2) The initializer then learns from the MCMC. Specifically, the initializer treats $\{(\tilde{Y}_i, C_i), i = 1,..., n\}$ produced by the MCMC as the training data. The key is that these $\{\tilde{Y}_i\}$ are obtained by the Langevin dynamics initialized from the $\{\hY_i, i = 1, ..., n\}$, which are generated by the initializer with {\em known} latent noise vectors $\{\hX_i, i = 1, ..., n\}$. Given $\{(\hX_i, \tilde{Y}_i, C_i), i = 1, ..., n\}$, we can learn $\alpha$ by minimizing $\frac{1}{n} \sum_{i=1}^{n} \Vert \tilde{Y}_i-g(\hX_i, C_i; \alpha)\Vert^2$, which is a nonlinear
regression of $\tilde{Y}_i$ on $(\hX_i, C_i)$ . This can be accomplished by gradient descent 
\begin{eqnarray} 
  \Delta \alpha \propto -  (\tilde{Y}_i-g(\hX_i, C_i; \alpha) \frac{\partial}{\partial \alpha}g(\hX_i, C_i; \alpha).  
 \label{eq:G}
 \end{eqnarray}
 
{\bf Mapping shift}: Initially $g(X, C; \alpha)$ maps $(\hX_i, C_i)$ to the initial solution $\hat{Y_i}$. After updating $\alpha$, $g(X, C; \alpha)$ maps $(\hX_i, C_i)$ to the refined solution $\tY_i$. Thus the updating of $\alpha$ absorbs the MCMC transitions that change $\hY_i$ to $\tY_i$. In other words, we distill the MCMC transitions of the refinement process into $g(X, C; \alpha)$.

Algorithm \ref{code:ccl} presents a description of the conditional learning with two models. See Figures \ref{fig:learn} and \ref{fig:learn1} for illustrations. 

Both computations can be carried out by back-propagation, and the whole algorithm is in the form of alternating back-propagation.

\begin{algorithm}
\caption{Cooperative conditional learning}
\label{code:ccl}
\begin{algorithmic}[1]
\Require
\Statex (1) training examples $\{(Y_i,C_i), i=1,...,n\}$
\Statex (2) numbers of Langevin steps $l$
\Statex (3) number of learning iterations $T$.
\Ensure
\Statex (1) learned parameters $\theta$ and $\alpha$,
\Statex (2) generated examples $\{\hY_i, \tY_i, i= 1, ..., n\}$.
\item[]
\State $t\leftarrow 0$, initialize $\theta$ and $\alpha$.
\Repeat 
\State {\bf Initialization by mapping}: For $i = 1, ..., n$, generate $\hX_i \sim \N(0, I_d)$, and generate  the initial solution $\hY_i = g(\hX_i, C_i; \alpha^{(t)}) + \epsilon_i$. 
\State {\bf Solve based on objective}: For $i = 1, ..., n$,  starting from $\hY_i$, run $l$ steps of Langevin dynamics to obtain the refined solution $\tY_i$,  each step following equation (\ref{eq:LangevinD}). 
\State {\bf Learn-objective by objective shift}: Update $\theta^{(t+1)} = \theta^{(t)} + \gamma_t L'(\theta^{(t)})$,  where $L'(\theta^{(t)})$ is computed according to (\ref{eq:learningD}). 
\State {\bf Learn-mapping by mapping shift}: Update $\alpha^{(t+1)} = \alpha^{(t)} + \gamma_t \Delta \alpha^{(t)} $,  where $\Delta \alpha^{(t)}$ is computed according to (\ref{eq:G})
\State Let $t \leftarrow t+1$
\Until $t = T$
\end{algorithmic}
\end{algorithm}

In Algorithm \ref{code:ccl}, the conditional energy- model is the primary model for conditional synthesis or recovery by MCMC sampling. The conditional generator model plays an assisting role to initialize the MCMC sampling. 

%
%

\section{Theoretical underpinning}
This section presents theoretical underpinnings of the model and the learning algorithms presented in the previous section. Readers who are more interested in applications and experiments can jump to the next section.

\subsection{Kullback-Leibler divergence} 

The Kullback-Leibler divergence between two distributions $p(x)$ and $q(x)$ is defined as ${\rm KL}(p\|q) = \E_{p}[\log (p(X)/q(X))]$. 

The Kullback-Leibler divergence between two conditional distributions $p(y|x)$ and $q(y|x)$ is defined as 
\begin{eqnarray}
    {\rm KL}(p\|q) &=& \E_p\left[\log \frac{p(Y|X)}{q(Y|X)}\right] \\
    &=& \int \log \frac{p(y|x)}{q(y|x)} p(x, y) dxdy, 
 \end{eqnarray}
 where the expectation is over the joint distribution $p(x, y) = p(x) p(y|x)$. 

\subsection{Slow thinking solver} 

The slow thinking solver model is
\begin{eqnarray} 
   p(Y|C; \theta) = \frac{p(Y,C;\theta)}{p(C;\theta)}
   =\frac{p(Y,C;\theta)} {\int p(Y,C; \theta) dY} \nonumber\\ 
   =\frac{1}{Z(C; \theta)} \exp \left[ f(Y,C; \theta)\right], \label{eq:D1}
\end{eqnarray}  
where 
\begin{eqnarray} 
    Z(C; \theta) = \int \exp \left[ f(Y,C; \theta)\right] dY
\end{eqnarray}
is the normalizing constant and is analytically intractable. 

Suppose the training examples $\{(Y_i, C_i), i = 1, ..., n\}$ are generated by the true joint distribution $f(Y, C)$, whose conditional distribution is $f(Y|C)$. 

For large sample $n \rightarrow \infty$, the maximum likelihood estimation of $\theta$ is to minimize the Kullback-Leibler divergence 
\begin{eqnarray}
\min_\theta  {\rm KL}(f(Y|C)\| p(Y|C; \theta)). 
\end{eqnarray} 
In practice, the expectation with respect to $f(Y, C)$ is approximated by the sample average. The difficulty with $ {\rm KL}(f(Y|C)\| p(Y|C; \theta))$ is that the $\log Z(C; \theta)$ term is analytically intractable, and its derivative has to be approximated by MCMC sampling from the model $p(Y|C; \theta)$. 

\subsection{Fast thinking initializer} 

The fast thinking initializer is 
\begin{eqnarray} 
  X \sim \N(0, I_d),  \;  Y = g(X, C; \alpha) + \epsilon,  \; \epsilon \sim \N(0, \sigma^2  I_D).   \label{eq:G1}
 \end{eqnarray}
 We use the notation $q(Y|C; \alpha)$ to denote the resulting conditional distribution. It is obtained by 
\begin{eqnarray} 
    q(Y|C; \alpha) = \int q(X) q(Y|X, C; \alpha) dX, 
 \end{eqnarray} 
 which is analytically intractable. 
 
 For large sample, the maximum likelihood estimation of $\alpha$ is to minimize the Kullback-Leibler divergence 
\begin{eqnarray}
\min_\alpha  {\rm KL}(f(Y|C)\| q(Y|C; \alpha)). 
\end{eqnarray} 
Again, the expectation with respect to $f(Y, C)$ is approximated by the sample average. The difficulty with ${\rm KL}(f(Y|C)\| q(Y|C; \alpha))$ is that $\log q(Y|C; \alpha)$ is analytically intractable, and its derivative has to be approximated by MCMC sampling of the posterior $q(X|Y, C; \alpha)$. 

\subsection{Objective shift: modified contrastive divergence} 

 Let $M(Y_1|Y_0, C; \theta)$ be the transition kernel of the finite-step MCMC that refines the initial solution $Y_0$ to the refined solution $Y_1$. Let $(M_\theta q)(Y_1|C; \alpha) = \int M(Y_1|Y_0, C; \theta)  q(Y_0|C; \alpha) dY_0$ be the distribution obtained by running the finite-step MCMC from $q(Y_0|C; \alpha)$. 
 
 Given the current initializer $q(Y|C; \alpha)$, the objective shift updates $\theta_t$ to $\theta_{t+1}$, and the update approximately follows the gradient of the following modified contrastive divergence \cite{Hinton2002a,xie2016cooperative}
 \begin{align}
 &{\rm KL}(f(Y|C)\| p(Y|C; \theta)) \nonumber \\- &{\rm KL}((M_{\theta_t} q)(Y|C; \alpha)\| p(Y|C; \theta)). \label{eq:MCD}
\end{align} 
Compare (\ref{eq:MCD}) with the MLE (\ref{eq:D1}),  (\ref{eq:MCD}) has the second divergence term $ {\rm KL}((M_{\theta_t} q)(Y|C; \alpha)\| p(Y|C; \theta))$ to cancel the $\log Z(C; \theta)$ term, so that its derivative is analytically tractable. The learning is to shift $p(Y|C; \theta)$ or its high value region around the mode from the refined solution provided by $(M_{\theta_t} q)(Y|C; \alpha)$ toward the observed solution given by $f(Y|C)$. If $(M_{\theta_t} q)(Y|C; \alpha)$ is close to $p(Y|C; \theta)$, then the second divergence is close to zero, and the learning is close to MLE update. 

\subsection{Mapping shift: distilling MCMC} 

Given the current solver model $p(Y|C; \theta)$, the mapping shift updates $\alpha_t$ to $\alpha_{t+1}$, and the update approximately follows the gradient of 
 \begin{eqnarray}
 {\rm KL}((M_\theta q)(Y|C; \alpha_{t})\| q(Y|C; \alpha)). \label{eq:DM}
\end{eqnarray} 
This update distills the MCMC transition $M_\theta$ into the model $q(Y|C; \alpha)$.  In the idealized case where the above divergence can be minimized to zero, then $q(Y|C; \alpha_{t+1}) = (M_\theta q)(Y|C; \alpha_{t})$. The limiting distribution of the MCMC transition $M_\theta$ is $p(Y|C; \theta)$,  thus the cumulative effect of the above update is to lead $q(Y|C; \alpha)$ close to $p(Y|C; \theta)$. 

Compare (\ref{eq:DM}) to the MLE (\ref{eq:G1}), the training data distribution becomes $(M_\theta q)(Y|C; \alpha_{t})$ instead of $f(Y|C)$. That is, $q(Y|C; \alpha)$ learns from how $M_\theta$ refines it. The learning is accomplished by mapping shift where the generated latent vector $X$ is known, thus does not need to be inferred (or the Langevin inference algorithm can initialize from the generated $X$). In contrast, if we are to learn from $f(Y|C)$, we need to infer the unknown $X$ by sampling from the posterior distribution. 

In the limit, if the algorithm converges to a fixed point, then the resulting $q(Y|C; \alpha)$ minimizes  ${\rm KL}((M_\theta q)(Y|C; \alpha)\| q(Y|C; \alpha))$, that is, $q(Y|C; \alpha)$ seeks to be the stationary distribution of the MCMC transition $M_\theta$, which is $p(Y|C; \theta)$. 

If the learned $q(Y|C; \alpha)$ is close to $p(Y|C; \theta)$, then $(M_{\theta_t} q)(Y|C; \alpha)$ is even closer to $p(Y|C; \theta)$. Then the learned $p(Y|C; \theta)$ is close to MLE because the second divergence term in (\ref{eq:MCD}) is close to zero.

\section{Experiments}
\textbf{Project page}: The code and more results can be found at 
\url{http://www.stat.ucla.edu/~jxie/CCoopNets/}

We test the proposed framework for conditional learning on a variety of vision tasks. According to the form of the conditional learning, we organize the experiments into two parts. In the first part (Experiment 1), we study conditional learning for a mapping from category (i.e., one-hot vector) to image, e.g., image generation conditioned on image class, while in the second part  (Experiment 2), we study conditional learning for a mapping from image to image, e.g., image-to-image translation. We propose a specific network architecture of our model in each experiment due to the different forms of input-output domains. 
Unlike the unconditioned cooperative learning framework \cite{xie2016cooperative, XieLuGao}, the conditioned framework needs to 
find a proper way to fuse the condition input $C$ into both the bottom-up ConvNet $f$ in the solver and the top-down ConvNet $g$ in the initializer, for the sake of capturing accurate conditioning information. An improper design can cause not only unrealistic but also condition-mismatched synthesized results.        

\subsection{Experiment 1: Category $\rightarrow$ Image}\label{exp:1}

\subsubsection{Network architecture}\label{exp:1_net}
We start form learning the conditional distribution of an image given a category or class label. The category information is encoded as a one-hot vector. The network architectures of the models in this experiment are given as follows.  

\begin{figure}[h]
\centering	
\includegraphics[width=.85\linewidth]{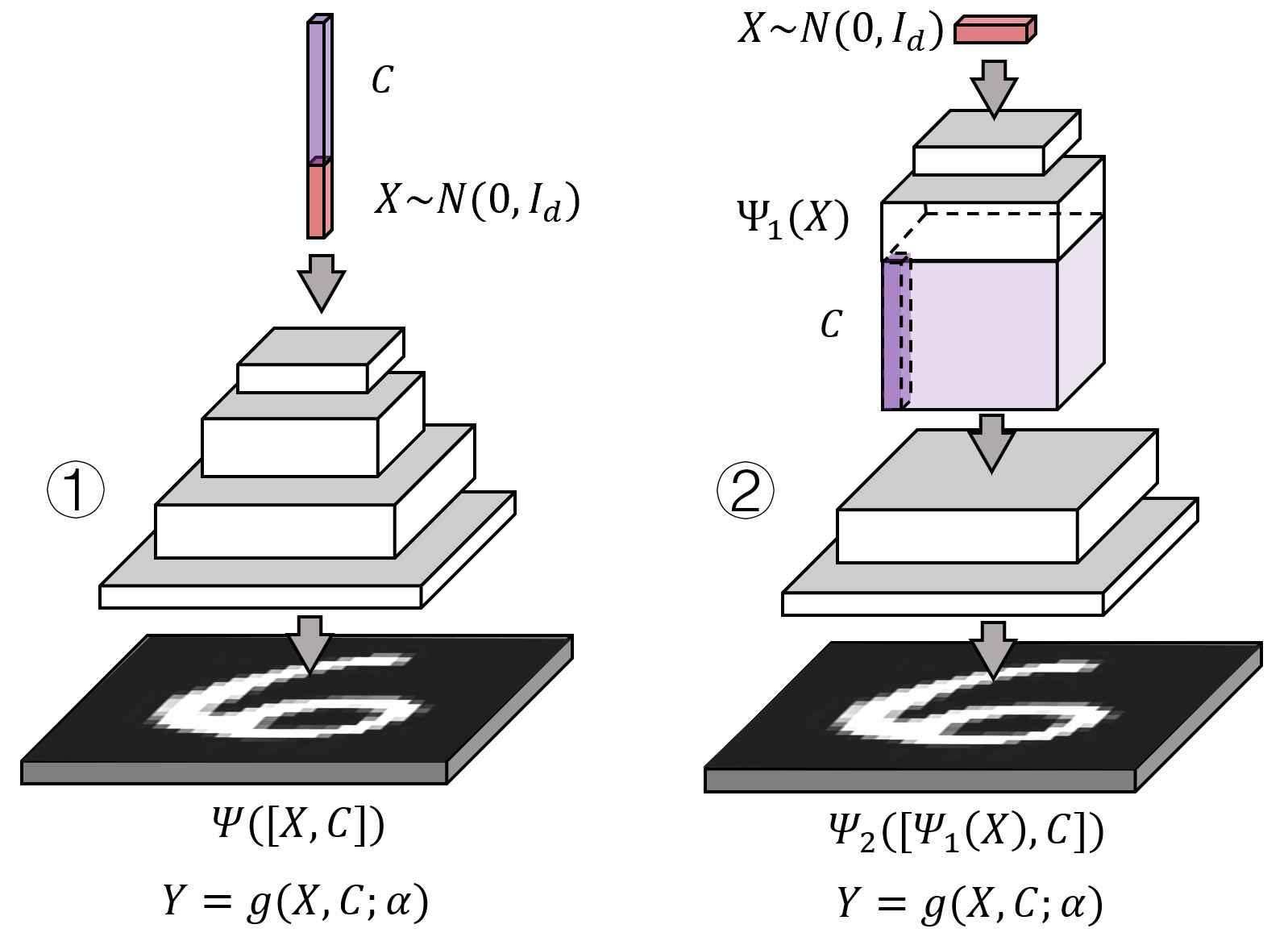}  
\caption{Network architecture of initializer (category-to-image synthesis). (1) early concatenation: a decoder $\Psi$ takes as input the concatenation of the condition vector $C$ and the latent noise vector $X \sim {\rm N}(0,I_d)$, and outputs an image $Y$. (2) late concatenation: a decoder takes as input only the latent noise vector $X \sim {\rm N}(0,I_d)$, and outputs an image $Y$, in which the condition $C$ is concatenated with the output of an intermediate layer. $\Psi_2$ is the sub-network after concatenation, while $\Psi_1$ is the sub-network before concatenation.}%
\label{structure_initializer}
\end{figure}

In the initializer, we can concatenate the one-hot vector $C$ with the latent noise vector $X$ sampled from ${\rm N}(0,I_d)$ as the input of the decoder $\Psi([X,C])$ to build a conditional generator $g(X,C;\alpha)$. The generator maps the input  into image $Y$ by several layers of deconvolutions. We call this setting ``early concatenation''. See Figure \ref{structure_initializer}(1) for an illustration. We can also adopt an architecture with ``late concatenation'', where the concatenation happens in the intermediate layer of the initializer. Specifically, we can first sample the latent noise vector $X$ from Gaussian noise prior ${\rm N}(0,I_d)$, and then decode $X$ to an intermediate result with spatial dimension $b \times b$ by a decoder $\Psi_1(X)$. The decoder consists of several layers of deconvolutions, each of which is followed by batch normalization \cite{ioffe2015batch} and  ReLU non-linear transformation. We then replicate the one-hot vector $C$ spatially and perform a channel concatenation with the intermediate output. After that, we generate the target image $Y$ from the concatenated result $[\Psi_1(X),C]$ by another decoder $\Psi_2([\Psi_1(X),C])$ that consists of several deconvolution layers. Batch normalization and ReLU layer are used between two consecutive deconvolution layers, and tanh non-linearity is added at the bottom layer. $g(X,C;\alpha)$ is the composition of $\Psi_1$ and $\Psi_2$. See Figure \ref{structure_initializer}(2) for an illustration. The details of the networks will be mentioned in the section of each experiment.

\begin{figure}[h]
\centering	
\includegraphics[width=.90\linewidth]{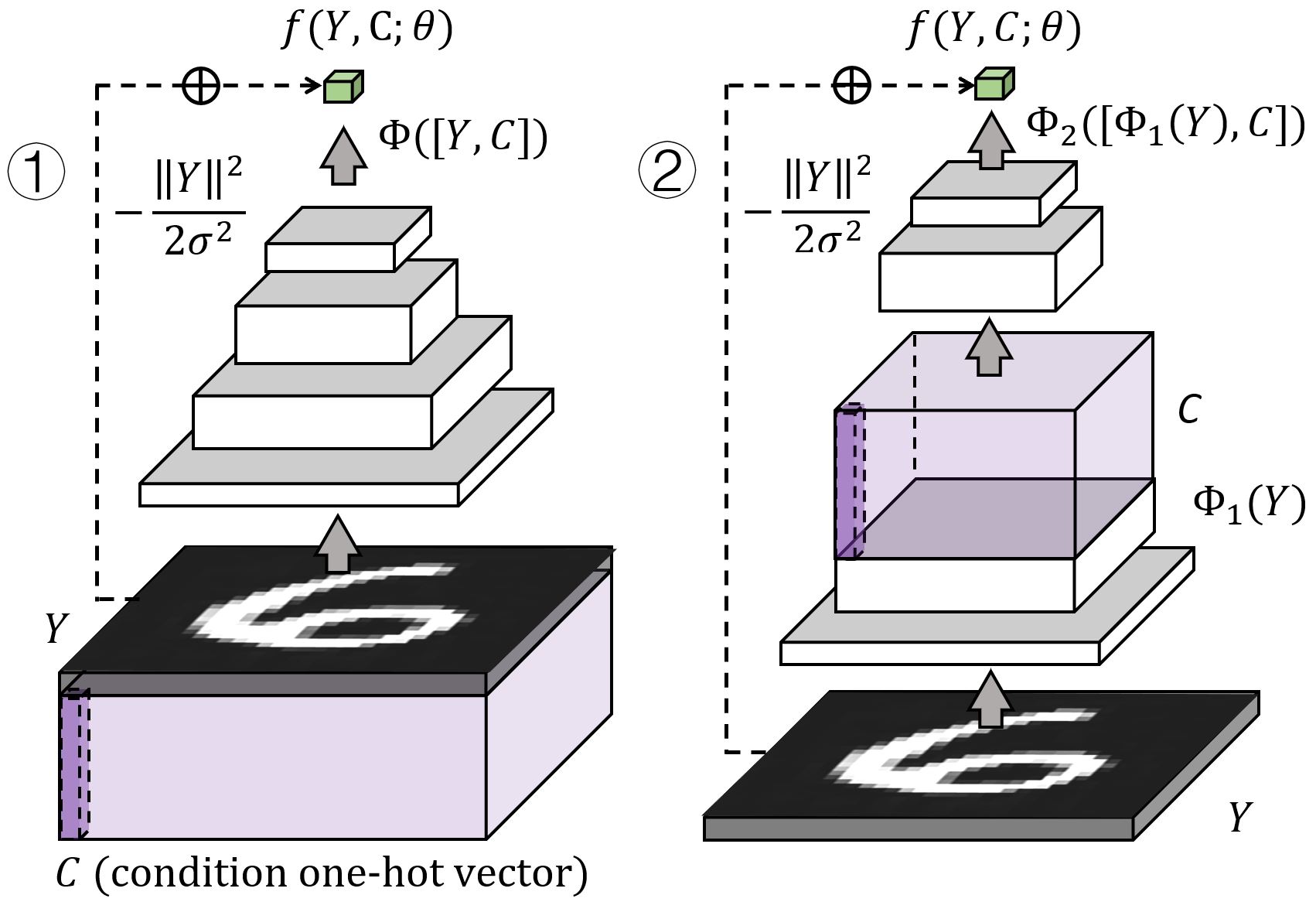}  
\caption{Network architecture of solver (category-to-image synthesis). (1) early concatenation: an encoder $\Phi$ takes as input the depth concatenation of the spatially replicated condition vector $C$ and the image $Y$, and outputs a scalar. The value function $f(Y,C;\theta)$ is defined as $\Phi([Y,C])-\Vert Y\Vert^2/2s^2$. (2) late concatenation: an encoder takes as input only the image $Y$, and outputs the negative energy, in which the condition $C$ is concatenated with the output of an intermediate layer. $\Phi_2$ is the sub-network after concatenation, while $\Phi_1$ is the sub-network before concatenation.}%
\label{structure_solver}
\end{figure}

To build the value function for the solver model, in the setting of ``early concatenation'', we first replicate the condition one-hot vector $C$ spatially and perform a depth concatenation with image $Y$, and then map them to a scalar by an encoder, $\Phi([Y,C])$, that consists of several layers of convolutions and ReLU non-linear transformations. The value function $f(Y,C;\theta)$ is designed as $\Phi([Y,C])-\Vert Y\Vert^2/2s^2$. This corresponds to an exponential tilting form in \cite{XieLuICML}, 
\begin{eqnarray}
   p(Y,C; \theta) = \frac{1}{Z(\theta)} \exp\left[ \Phi(Y,C; \theta)\right]p_0(Y), 
\end{eqnarray}
where $p_0(Y)$ is Gaussian white noise distribution, i.e., $p_0(Y) \propto \exp (-\Vert Y\Vert^2/2s^2)$, and $s$ is a hyperparameter for the standard deviation of $p_0$. See Figure \ref{structure_solver}(1) for an illustration. As to the ``late concatenation'', we first encode the image $Y$ to an intermediate result with spatial dimension $a \times a$ by an encoder $\Phi_1(Y)$, which consists of several layers of convolutions and ReLU non-linear transformations, and then we replicate the one-hot vector $C$ spatially and perform a depth concatenation with the intermediate result. The value function is defined by another encoder $\Phi_2([\Phi_1(Y), C])$ plus $-\Vert Y\Vert^2/2s^2$, in which the encoder takes as input the concatenated result $[\Phi_1(Y), C]$ and outputs a scalar by performing several layers of convolutions and ReLU non-linear transformations. See Figure \ref{structure_solver}(2) for an illustration. Detailed network configuration will be discussed in the section of each experiment.  

\subsubsection{Conditional image generation on grayscale images} \label{sec:mnist}
 
We first test our model on two grayscale image datasets, such as MNIST \cite{lecun1998gradient} and fashion-MNIST \cite{xiao2017fashion}. The former is a dataset of handwritten digit images, and the latter is a dataset of fashion product images. Each of them consists of 70,000 $28 \times 28$ images, each of which is associated with a label from 10 classes. In each dataset, 60,000 examples are used for training and the rest are for testing. We learn our model on each of them respectively, conditioned on their class labels that are encoded as one-hot vectors. Since these two datasets are similar in number of classes, image size, data size, and image format (i.e., grayscale), we use the same model for them.

We adopt the setting of ``early concantenation'' introduced in section \ref{exp:1_net} for the initializer. To be specific, $g(X,C; \alpha)$ is a generator that maps the $1 \times 1 \times 138$ concatenated result (Note that the dimension of $X$ is 128, and the size of $C$ is 10.) to a $28 \times 28 $ grayscale image by 4 layers of deconvolutions with kernel sizes $\{4,4, 4, 4\}$, up-sampling factors $\{1,2, 2, 2\}$ and numbers of output channels $\{256,128,64,1\}$ at different layers. The last deconvolution layer is followed by a tanh operation, and each of the others is followed by batch normalization and ReLU operation.

We adopt the setting of ``late concatenation'' introduced in section \ref{exp:1_net} for the solver. Specifically, $\Phi_1(Y)$ consists of 2 layers of convolutions with filter sizes $\{5,3\}$, down-sampling factors $\{2,2\}$ and numbers of output channels $\{64,128\}$. The concatenated output is of size $7 \times 7 \times 138$. (Note that the number of the output channels of $\Phi_1$ is 128, and the size of $C$ is 10.) $\Phi_2([\Phi_1(Y),C])$ is a 2-layer ConvNet, where the first layer has 256 $3 \times 3$ filters, and the last layer is a fully-connected layer with 100 filters.

We use Adam \cite{kingma2014adam} to optimize the solver with initial learning rate 0.0008, $\beta_1 = 0.5$ and $\beta_2 = 0.999$, and the initializer with initial learning rate 0.0001, $\beta_1 = 0.5$ and $\beta_2 = 0.999$. The mini-batch size is 300. The number of paralleled MCMC chains is 300. The number of Langevin dynamics steps is $l=16$. The step size $\delta$ of Langevin dynamics is 0.0008. The standard deviation of the residual in the initializer is $\sigma=0.3$, and the standard deviation $s$ of the reference distribution $p_0$ in the solver is 0.016. We run 1,600 epochs to train the model, where we disable the noise term in Langevin dynamics after the first 100 epochs. 

\begin{figure}[h]
\centering	
\includegraphics[width=.65\linewidth]{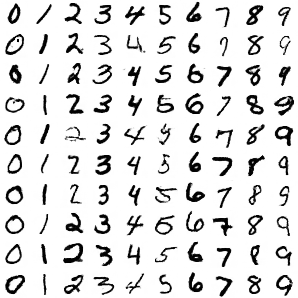}  
\caption{Generated MNIST handwritten digits. Each column is conditioned on one class label and each row is a different synthesized sample. The size of the generated images is $28 \times 28$.}%
\label{syn_digit}
\end{figure}

\begin{figure}[h]
\centering	
\includegraphics[width=.65\linewidth]{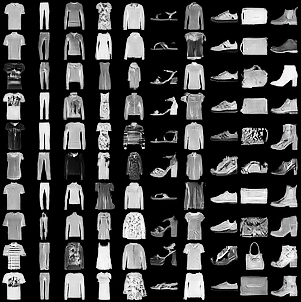}  
\caption{Generated fashion MNIST images. Each column is conditioned on one class label and each row is a different synthesized sample. The size of the generated images is $28 \times 28$.}%
\label{syn_fashion}
\end{figure}

Figure \ref{syn_digit} shows some of the generated samples conditioned on the class labels after training on the MNIST dataset. Each column is conditioned on one label and each row is a different generated sample. Figure \ref{syn_fashion} shows the results for the fashion-MNIST dataset. The qualitative results show that our method can learn realistic conditional models.

To quantitatively evaluate the learned conditional distribution, we use ``Fréchet Inception Distance'' \cite{heusel2017gans} (FID) score as a metric to measure
the dissimilarity between the distributions of the observed and the synthesized examples. Specifically, we compute the distance between feature vectors extracted from observed and synthesized examples by a pre-trained Inception  model \cite{szegedy2016rethinking}, with
the following formula
\begin{eqnarray} 
\text{FID}=||\tilde{\mu} - \mu ||^2 + \text{Tr}\left(  \tilde{\Sigma} + \Sigma - 2(\tilde{\Sigma} \Sigma)^{1/2}\right), \nonumber 
\end{eqnarray} 
where $V \sim \N(\mu, \Sigma)$ and $\tilde{V} \sim \N(\tilde{\mu}, \tilde{\Sigma})$ are the 2,048-dimensional feature vectors of the observed and synthesized examples, respectively. They are the outputs taken as the activations from the global spatial pooling layer of the Inception model. We can fit a multi-variate Gaussian to feature vectors $\{V_i\}$ and $\{\tilde{V}_i\}$ separately, to obtain means $\mu, \tilde{\mu}$ and variances $\Sigma,\tilde{\Sigma}$ for the observed and synthesized distributions respectively. A lower FID score implies better qualities of the synthesized images. 

To compute FID score, we sample 10,000 examples from the learned conditional distribution by first sampling the class label $C$ from the uniform prior distribution, and $X$ from ${\rm N}(0,I_d)$, then the initializer and the solver model cooperatively generate the synthesized example from the sampled $C$ and $X$. Table \ref{MNIST_FID} shows a comparison of FID scores of different methods on two datasets. Our method achieves better results than other conditional and unconditional baseline methods in terms of generation quality evaluated by FID. Those baselines include GAN-based, flow-based, and variational inference methods. 

\begin{figure*}[h]
\centering	
\begin{tabular}{cccccc} 
\includegraphics[width=.14\linewidth]{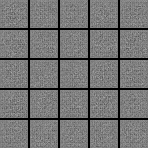}  &
\includegraphics[width=.14\linewidth]{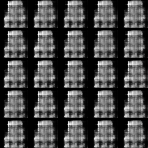} &
\includegraphics[width=.14\linewidth]{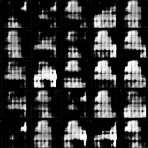} &
\includegraphics[width=.14\linewidth]{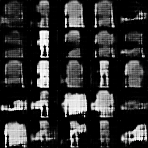} &
\includegraphics[width=.14\linewidth]{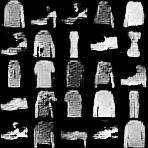}&
\includegraphics[width=.14\linewidth]{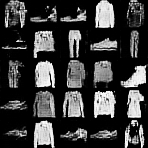}  \\
$t=0$ (325.78)& $t=10$ (308.87)&$t=20$ (239.74)&$t=30$ (139.48)&$t=40$ (80.33)&$t=50$ (73.48)\\
\includegraphics[width=.14\linewidth]{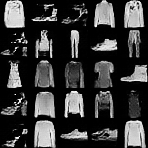} &
\includegraphics[width=.14\linewidth]{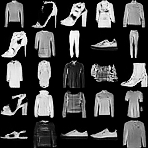} &
\includegraphics[width=.14\linewidth]{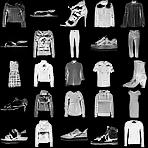} &
\includegraphics[width=.14\linewidth]{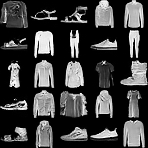} &
\includegraphics[width=.14\linewidth]{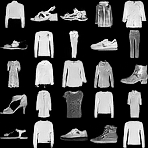} &
\includegraphics[width=.14\linewidth]{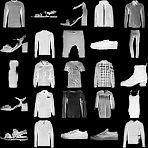} \\
$t=100$ (32.61)& $t=300$ (10.47)&$t=500$ (10.08)&$t=800$ (8.91)&$t=1000$ (8.43)&$t=1280$ (8.20)\\
\end{tabular}
\caption{Image generation by the models at different training epochs. For each epoch $t$, 25 examples of synthesized images are displayed. The numbers in parentheses are the corresponding FID scores that reflect the qualities of the synthesized images. The images at the same position of image matrix of different training epochs are generated from the same condition.}%
\label{fig:synthesis_sequence}
\end{figure*}

Figure \ref{fig:synthesis_sequence} displays some examples of the synthesized images at different training epochs along with the corresponding FID scores. The images shown are generated by the solver. The images at the same position of $5 \times 5$ image matrix of different training epochs share the same condition $C$, i.e., the class label. We can find that as the cooperative training progresses, the synthesized images become more and more realistic and the FID scores become lower and lower. Additionally, the learned connection between the condition (i.e., class label) and the target (i.e., image) becomes more and more accurate in the sense that when the model converges, even though the appearances of the synthesized images vary at different epochs, they are always consistent with their input conditions.

\begin{table}[h]
\centering
\begin{small}
\caption{The Fréchet Inception Distance (FID) scores of different models trained on MNIST and fashion-MNIST datasets, the smaller the FID, the better the performance.}
\label{MNIST_FID}
\begin{tabular}{|c|lcc|} \hline
& Model & MNIST & fashion-MNIST\\ \hline \hline
\multirow{10}{*}{\rotatebox{90}{\footnotesize unconditional}} & GLO  \cite{bojanowski2017optimizing} & 49.60 & 57.70\\
& VAE \cite{kingma2013auto} & 21.85 & 69.84\\
& BEGAN \cite{berthelot2017began}  & 13.54 & 15.90\\
&EBGAN \cite{zhao2016energy} & 11.10 & 41.32\\
& GLANN \cite{hoshen2019non} & 8.60 & 13.10\\
& WGAN \cite{arjovsky2017wasserstein} & 7.07 & 28.17\\
& LSGAN \cite{mao2017least} & 6.75 &14.72\\
& DCGAN \cite{radford2015unsupervised} & 4.54 &8.22\\
& InfoGAN \cite{chen2016infogan} & 28.09  & -\\
& GLF \cite{xiao2019generative} & 5.80  & 10.30 \\ 
\hline
\multirow{8}{*}{\rotatebox{90}{\footnotesize conditional}} & CGlow \cite{liu2019conditional} & 29.64 & - \\
& CAGlow \cite{liu2019conditional} & 26.34 & -  \\
& VCGAN \cite{hu2019variational} & - & 13.8 \\ 
& CVAE-GAN \cite{bao2017cvae} & - & 15.9 \\
&CVAE \cite{sohn2015learning} & 20.00 & 36.64 \\ 
&ACGAN \cite{odena2017conditional} & 12.55 & 49.11\\ 
&CGAN \cite{mirza2014conditional} & 5.91 & 11.92\\
&\textbf{CCoopNets (ours)} & \textbf{4.50} & \textbf{8.20}\\
\hline
\end{tabular}
\end{small}
\end{table}


\begin{figure*}[h]
\centering	
\begin{tabular}{ccc}
\includegraphics[width=.3\linewidth]{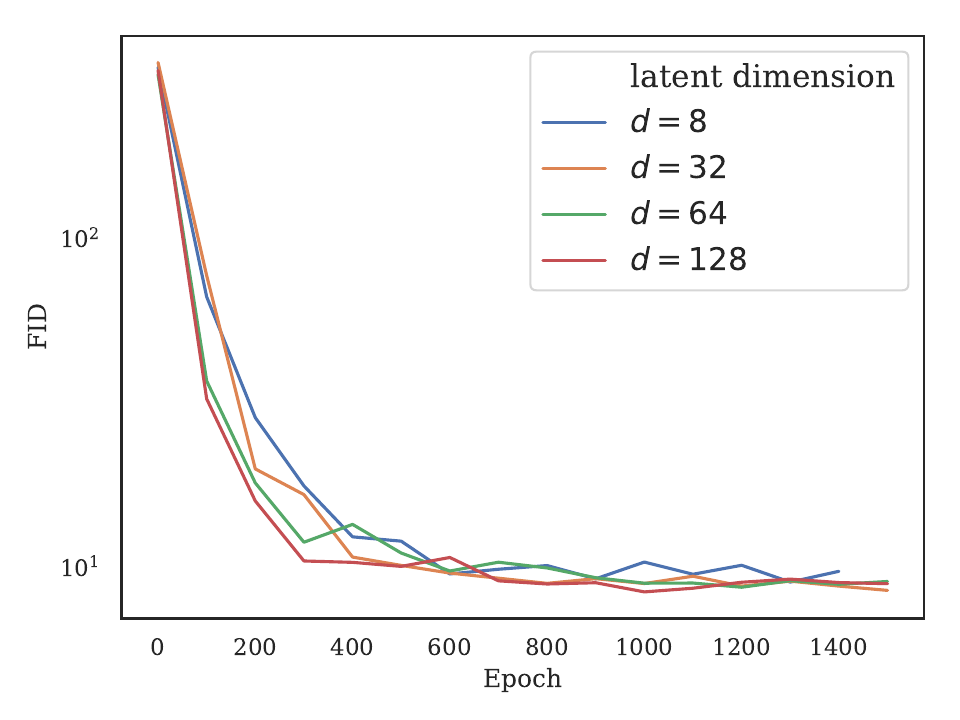}  &
\includegraphics[width=.3\linewidth]{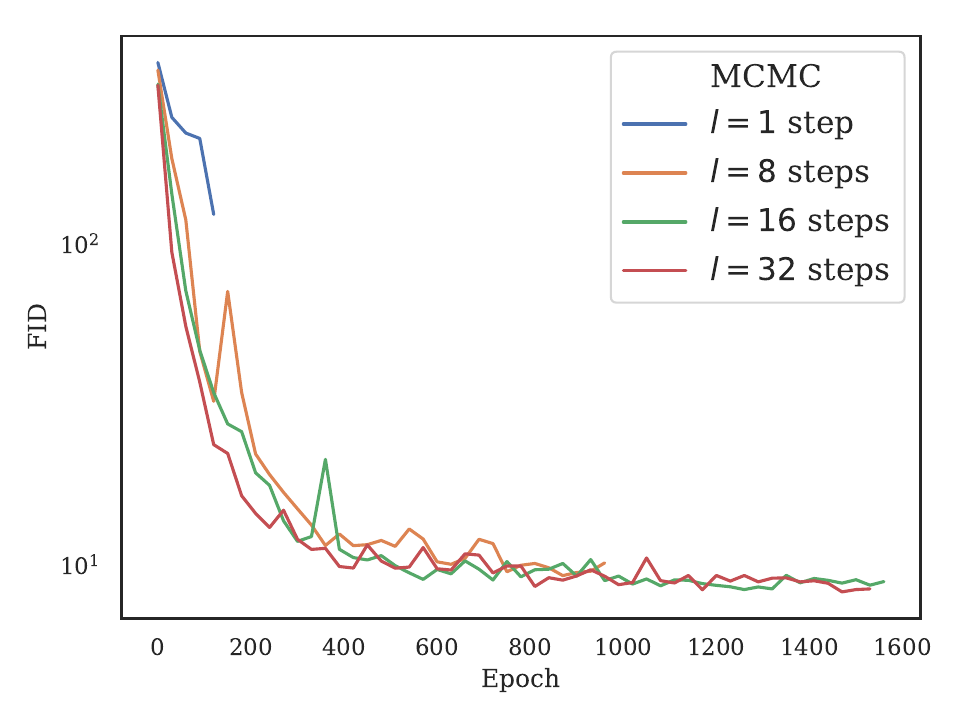} &
\includegraphics[width=.3\linewidth]{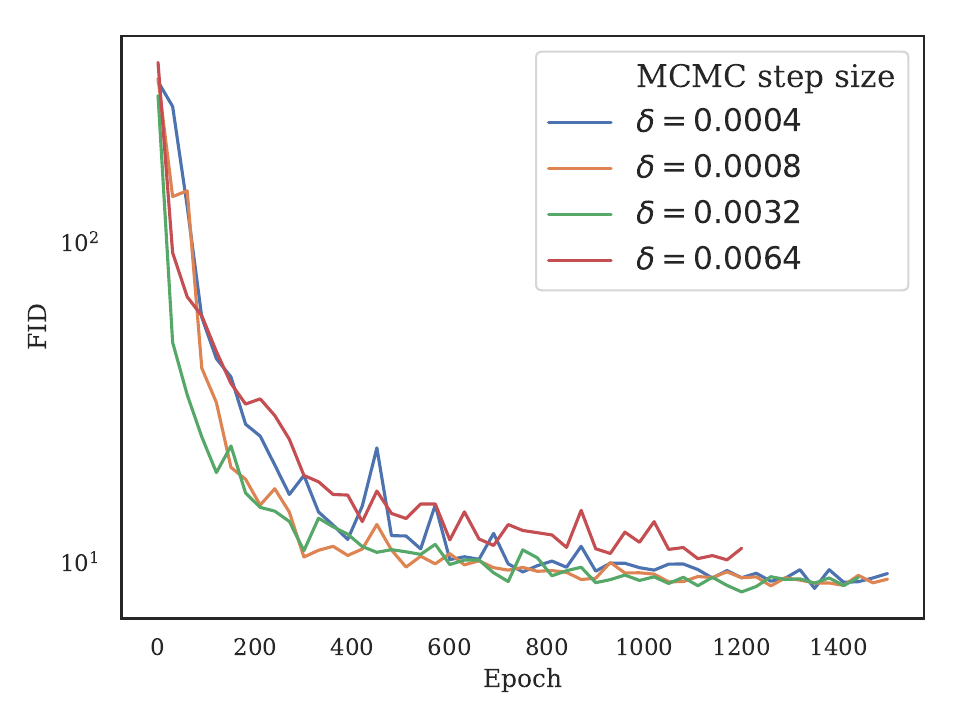} \\
(a) number of latent dimension & (b) number of Langevin steps & (c) step size of Langevin refinement
\end{tabular}
\caption{Model analysis on fashion-MNIST dataset. (a) Influence of the number of latent dimension $d$ of the fast-thinking initializer. We set $l=16$ and $\delta=0.0008$. (b) Influence of the number $l$ of Langevin refinement steps by the slow-thinking solver. We set $d=64$ and $\delta=0.0008$. (c) Influence of the step size $\delta$ of Langevin refinement of the slow-thinking solver. We set $d=128$ and $l=16$.}%
\label{fig:z_plot}
\end{figure*}

We study the influences of different choices of some hyper-parameters, such as the number of dimension $d$ of the latent space $X$ in the initializer, the number of Langevin refinement steps $l$, and the step size $\delta$ of each Langevin. Figure \ref{fig:z_plot} depicts the influences of varying $d$, $l$ and $\delta$, respectively, while training on fashion-MNIST dataset. Each curve represents the testing FID scores over training epochs. We observe that (1) the quality of synthesis decreases with decreasing $d$. (2) the more the number of MCMC refinement steps, the stabler the
learning process, and the more time-consuming the refinement process of the solver. With a small $l$, e.g., 1 or 8, the cooperative learning tends to fail easily at the early  stage of training because the slow-thinking solver distills an insufficient refinement process to the initializer such that the latter can not provide good initial solutions for the former. Figure \ref{fig:z_plot}(b) shows that the learning curves for $l=1$ (in blue) and $l=8$ (in orange) are terminated early due to  failures occurred during training. Table \ref{fashion_MNIST_time} shows a comparison of computational time per epoch with different numbers of Langevin steps $l$ and different numbers of latent dimensions $d$. A choice of $l=16$ or $32$ appears reasonable. The influence of $d$ on the computational time is not significant. (3) A large Langevin step size allows the model to learn faster to generate high quality images, at the cost of arriving on a sub-optimal synthesis of images. A smaller Langevin step size may allow the model to generate more realistic images but it may take more Langevin steps.

\begin{table}[h]
\centering
\begin{small}
\caption{Comparison of computational time (in seconds) per epoch with different numbers of Langevin refinement steps and different numbers of latent dimensions for class-conditioned image generation on fashion-MNIST dataset. The running times were recorded in a PC with an Intel i7-6700k CPU and a Titan Xp GPU.}
\label{fashion_MNIST_time}
\begin{tabular}{|l|ccccc|} \hline
 & $l=1$  & $l=8$ & $l=16$ & $l=32$ & $l=64$\\ \hline \hline
  $d=8$   & 8.98 & 20.38 & 26.88 & 46.74 & 86.93\\
   $d=32$  & 9.23 & 20.21 & 27.04 & 46.95 & 86.95  \\
 $d=64$   & 9.12 & 20.10 & 27.55 & 47.22  & 87.06\\
  $d=128$   & 9.37 & 20.50 & 27.76 & 48.62 & 86.92\\
\hline
\end{tabular}
\end{small}
\end{table}

\subsubsection{Conditional image generation on Cifar-10}

We also test the proposed framework on Cifar-10 \cite{krizhevsky2009learning} object dataset, which contains 10-class 60,000 training images of $32 \times 32$ pixels. Compared with the MNIST dataset, Cifar-10 contains training images with more complicated visual patterns. 

As to the initializer, we adopt the ``late concatenation'' setting. Specifically, $\Psi_1(X)$ is a decoder that maps 100-dimensional $X$ (i.e., $1 \times 1 \times 100$) to an intermediate output with spatial dimension $8 \times 8$ by 2 layers of deconvolutions with kernel sizes $\{4,5\}$, up-sampling factors $\{1,2\}$ and numbers of output channels $\{256,128\}$ at different layers from top to bottom, respectively. The condition $C$ is a 10-dimensional one-hot vector to represent the class. $\Psi_2([\Psi_1(X),C])$ is a generator that maps the $8 \times 8 \times 138$ concatenated result to a $32 \times 32 \times 3$ image by 2 layers of deconvolutions with kernel sizes $\{5,5\}$, up-sampling factors $\{2,2\}$ and numbers of output channels $\{64,3\}$ at different layers.

We adopt the ``late concatenation'' setting for the solver. $\Phi_1(Y)$ consists of 2 layers of convolutions with filter sizes $\{5,3\}$, down-sampling factors $\{2,2\}$ and numbers of output channels $\{64,128\}$. The concatenated output is of size $8 \times 8 \times 138$. $\Phi_2([\Phi_1(Y), C])$ is a 2-layer bottom-up ConvNet, where the first layer has 256 $3 \times 3$ filters, and the last layer is a fully connected layer with 100 filters.

We use the Adam for optimization. The initial learning rates for the solver and initializer are 0.002 and 0.0064, respectively. The joint models are trained with mini-batches of size 300. The number of paralleled MCMC chains is also 300. The number of Langevin dynamics steps is 8. The step size $\delta$ of Langevin dynamics is 0.0008. We run 2,000 epochs to train the model, where we disable the noise term in Langevin dynamics in the last 1,500 ones.

\begin{figure}[h]
\centering	
\includegraphics[width=.07\linewidth]{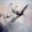}
\includegraphics[width=.07\linewidth]{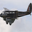} \hspace{2mm}
\includegraphics[width=.07\linewidth]{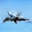} 
\includegraphics[width=.07\linewidth]{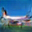} 
\includegraphics[width=.07\linewidth]{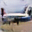} 
\includegraphics[width=.07\linewidth]{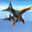} 
\includegraphics[width=.07\linewidth]{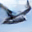} 
\includegraphics[width=.07\linewidth]{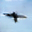} 
\includegraphics[width=.07\linewidth]{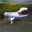} 
\includegraphics[width=.07\linewidth]{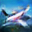} 
\\ \vspace{1mm}

\includegraphics[width=.07\linewidth]{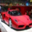}
\includegraphics[width=.07\linewidth]{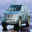} \hspace{2mm}
\includegraphics[width=.07\linewidth]{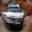}
\includegraphics[width=.07\linewidth]{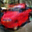}
\includegraphics[width=.07\linewidth]{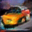}
\includegraphics[width=.07\linewidth]{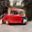}
\includegraphics[width=.07\linewidth]{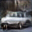}
\includegraphics[width=.07\linewidth]{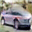}
\includegraphics[width=.07\linewidth]{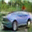}
\includegraphics[width=.07\linewidth]{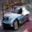}\\ \vspace{1mm}

\includegraphics[width=.07\linewidth]{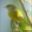}
\includegraphics[width=.07\linewidth]{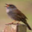} \hspace{2mm}
\includegraphics[width=.07\linewidth]{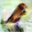}
\includegraphics[width=.07\linewidth]{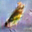}
\includegraphics[width=.07\linewidth]{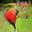}
\includegraphics[width=.07\linewidth]{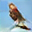}
\includegraphics[width=.07\linewidth]{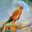}
\includegraphics[width=.07\linewidth]{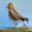}
\includegraphics[width=.07\linewidth]{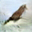}
\includegraphics[width=.07\linewidth]{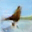}
\\ \vspace{1mm}

\includegraphics[width=.07\linewidth]{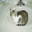}
\includegraphics[width=.07\linewidth]{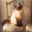} \hspace{2mm}
\includegraphics[width=.07\linewidth]{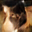}
\includegraphics[width=.07\linewidth]{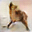}
\includegraphics[width=.07\linewidth]{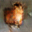}
\includegraphics[width=.07\linewidth]{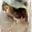}
\includegraphics[width=.07\linewidth]{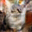}
\includegraphics[width=.07\linewidth]{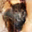}
\includegraphics[width=.07\linewidth]{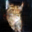}
\includegraphics[width=.07\linewidth]{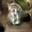}\\ \vspace{1mm}

\includegraphics[width=.07\linewidth]{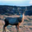}
\includegraphics[width=.07\linewidth]{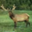} \hspace{2mm}
\includegraphics[width=.07\linewidth]{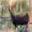}
\includegraphics[width=.07\linewidth]{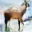}
\includegraphics[width=.07\linewidth]{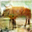}
\includegraphics[width=.07\linewidth]{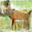}
\includegraphics[width=.07\linewidth]{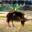}
\includegraphics[width=.07\linewidth]{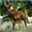}
\includegraphics[width=.07\linewidth]{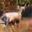}
\includegraphics[width=.07\linewidth]{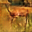}  \\ \vspace{1mm}

\includegraphics[width=.07\linewidth]{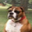}
\includegraphics[width=.07\linewidth]{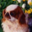} \hspace{2mm}
\includegraphics[width=.07\linewidth]{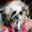}
\includegraphics[width=.07\linewidth]{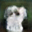}
\includegraphics[width=.07\linewidth]{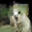}
\includegraphics[width=.07\linewidth]{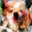}
\includegraphics[width=.07\linewidth]{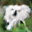}
\includegraphics[width=.07\linewidth]{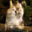}
\includegraphics[width=.07\linewidth]{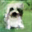}
\includegraphics[width=.07\linewidth]{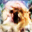}  \\ \vspace{1mm}

\includegraphics[width=.07\linewidth]{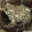}
\includegraphics[width=.07\linewidth]{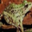} \hspace{2mm}
\includegraphics[width=.07\linewidth]{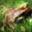} 
\includegraphics[width=.07\linewidth]{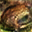}
\includegraphics[width=.07\linewidth]{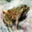}
\includegraphics[width=.07\linewidth]{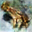}
\includegraphics[width=.07\linewidth]{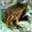}
\includegraphics[width=.07\linewidth]{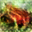}
\includegraphics[width=.07\linewidth]{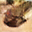}
\includegraphics[width=.07\linewidth]{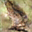}  \\ \vspace{1mm}

\includegraphics[width=.07\linewidth]{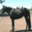}
\includegraphics[width=.07\linewidth]{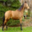} \hspace{2mm}
\includegraphics[width=.07\linewidth]{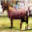}
\includegraphics[width=.07\linewidth]{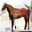}
\includegraphics[width=.07\linewidth]{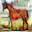}
\includegraphics[width=.07\linewidth]{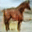}
\includegraphics[width=.07\linewidth]{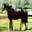}
\includegraphics[width=.07\linewidth]{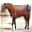}
\includegraphics[width=.07\linewidth]{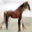}
\includegraphics[width=.07\linewidth]{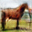}\\ \vspace{1mm}

\includegraphics[width=.07\linewidth]{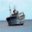}
\includegraphics[width=.07\linewidth]{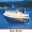} \hspace{2mm}
\includegraphics[width=.07\linewidth]{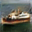}
\includegraphics[width=.07\linewidth]{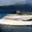}
\includegraphics[width=.07\linewidth]{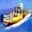}
\includegraphics[width=.07\linewidth]{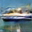}
\includegraphics[width=.07\linewidth]{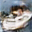}
\includegraphics[width=.07\linewidth]{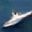}
\includegraphics[width=.07\linewidth]{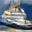}
\includegraphics[width=.07\linewidth]{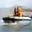}\\ \vspace{1mm}

\includegraphics[width=.07\linewidth]{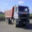}
\includegraphics[width=.07\linewidth]{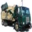} \hspace{2mm}
\includegraphics[width=.07\linewidth]{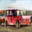}
\includegraphics[width=.07\linewidth]{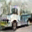}
\includegraphics[width=.07\linewidth]{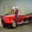}
\includegraphics[width=.07\linewidth]{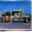}
\includegraphics[width=.07\linewidth]{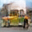}
\includegraphics[width=.07\linewidth]{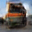}
\includegraphics[width=.07\linewidth]{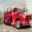}
\includegraphics[width=.07\linewidth]{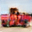}
\caption{Generated Cifar-10 object images. Each row is conditioned on one category label. The first two columns are training images, and the remaining columns display generated images conditioned on their labels. The image size is $32 \times 32$ pixels. The categories are airplane, automobile, bird, cat, deer, dog, frog, horse, ship, and truck from top to bottom.}%
\label{syn_cifar}
\end{figure}

Figure \ref{syn_cifar} shows the generated object patterns. Each row is conditioned on one category. The first two columns display some typical training examples, while the rest columns show generated images conditioned on labels. We evaluate the learned conditional distribution by computing the inception scores of the generated examples. Table \ref{tab:cifar} compares our framework against two baselines, which are two conditional models based on GANs. The proposed model performs better than the baselines. We also found that in the proposed method, the solution provided by the initializer is indeed further refined by the solver in terms of inception score.

\begin{table}
\centering
\begin{small}
\caption{Inception scores of different models trained on Cifar-10 dataset. The larger the inception score, the better the performance.}
\label{tab:cifar}
\begin{tabular}{|c|lc|} \hline
& Model & Inception score \\ \hline \hline

\multirow{5}{*}{\rotatebox{90}{\footnotesize unconditional}} & PixelCNN \cite{oord2016pixel} & 4.60  \\
& PixelIQN \cite{ostrovski2018autoregressive} & 5.29  \\
& DCGAN \cite{radford2015unsupervised} & 6.40  \\
& WGAN-GP \cite{gulrajani2017improved} & 6.50  \\ 
& ALI \cite{dumoulin2016adversarially} & 5.34 \\
\hline

\multirow{4}{*}{\rotatebox{90}{\footnotesize conditional}} & CGAN \cite{salimans2016improved} & 6.58\\ 
& Conditional SteinGAN   \cite{wang2016learning}
& 6.35 \\ 
& initializer (ours)  &  6.63 \\ 
& solver (ours)  &  \textbf{7.30}  \\ \hline
\end{tabular}
\end{small}
\end{table}

\subsubsection{Disentangling style and category}
To test the inference power of the fast-thinking initializer, which is trained jointly with the slow-thinking solver, we apply the learned initializer to a task of style transfer from an unseen testing image in one caegory  onto other categories. The models are first trained on SVHN \cite{netzer2011reading} dataset that contains 10 classes of digits collected from street view house numbers. The network architectures of initializer and solver are similar to those used in Section \ref{sec:mnist}, except that the training images in this experiment are RGB images and they are of size $32 \times 32$ pixels. With the learned initializer, we first infer the latent variables $X$ corresponding to that testing image. We then fix the inferred latent vector, change the category label $C$, and generate the different categories of images with the same style as the testing image by the learned model. Given a testing image $Y$ with known category label $C$, the inference of the latent vector $X$ can be performed by directly sampling from the posterior distribution $p(X|Y, C; \alpha)$ via Langevin dynamics, which iterates
\begin{eqnarray} 
\begin{aligned}
 & X_{\tau+1}=X_{\tau} + sU_{\tau} +\\
  & \frac{s^2}{2} \left[ \frac{1}{\sigma^2} (Y-g(X_{\tau}, C; \alpha)) \frac{\partial}{\partial X}g(X_{\tau}, C; \alpha) - X_{\tau} \right]. 
\end{aligned}
 \label{eq:styleInfer}
 \end{eqnarray}
 
\begin{figure}[h]
\centering
\includegraphics[height=.071\linewidth]{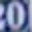} \hspace{2mm}
\includegraphics[height=.071\linewidth]{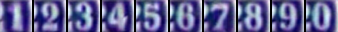} \\
\vspace{1mm}
\includegraphics[height=.071\linewidth]{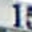} \hspace{2mm}
\includegraphics[height=.071\linewidth]{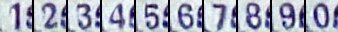} \\
\vspace{1mm}
\includegraphics[height=.071\linewidth]{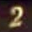} \hspace{2mm}
\includegraphics[height=.071\linewidth]{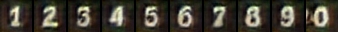} \\
\vspace{1mm}

\includegraphics[height=.071\linewidth]{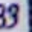} \hspace{2mm}
\includegraphics[height=.071\linewidth]{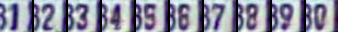} \\
\vspace{1mm}
\includegraphics[height=.071\linewidth]{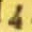} \hspace{2mm}
\includegraphics[height=.071\linewidth]{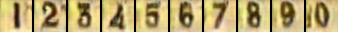} \\
\vspace{1mm}

\includegraphics[height=.071\linewidth]{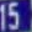} \hspace{2mm}
\includegraphics[height=.071\linewidth]{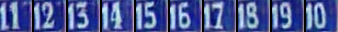} \\
\vspace{1mm}
\includegraphics[height=.071\linewidth]{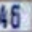} \hspace{2mm}
\includegraphics[height=.071\linewidth]{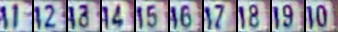}\\
\vspace{1mm}

\includegraphics[height=.071\linewidth]{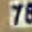} \hspace{2mm}
\includegraphics[height=.071\linewidth]{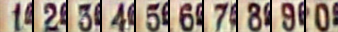} \\
\vspace{1mm}
\includegraphics[height=.071\linewidth]{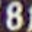} \hspace{2mm}
\includegraphics[height=.071\linewidth]{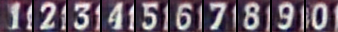} \\
\vspace{1mm}
\includegraphics[height=.071\linewidth]{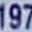} \hspace{2mm}
\includegraphics[height=.071\linewidth]{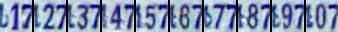} \\
\caption{Style transfer. The trained initializer can disentangle the style and the category such that the style information can be inferred from a testing image and transferred to other categories. The first column shows testing images. The other columns show style transfer by the model, where the style latent variable $X$ of each row is set to the value inferred from the testing image in the first column by the Langevin inference. Each column corresponds to a different category label $C$.}
\label{fig:Style_transfer}
\end{figure}

If the category label of the testing image is unknown, we need to infer both $C$ and $X$ from $Y$. Since $C$ is a one-hot vector, in order to adopt a gradient-based method to infer $C$, we adopt a continuous approximation by reparametrizing $C$ using a softMax transformation on the auxiliary continuous variables $A$. Specifically, let $C = (c_k, k = 1, ..., K)$ and $A = (a_k, k = 1, ..., K)$, we reparametrize $C=v(A)$ where $c_k = \exp(a_k)/\sum_k'  \exp(a_k')$, for $k = 1, ..., K$, 
and assume the prior for $A$ to be ${\rm N}(0, I_K)$. Then the Langevin dynamics for sampling $A \sim p(A|Y,X)$ iterates
 \begin{eqnarray} 
\begin{aligned}
 & A_{\tau+1}=A_{\tau} + sU_{\tau} +\\
  & \frac{s^2}{2} \left[ \frac{1}{\sigma^2} (Y-g(X_{\tau}, v(A); \alpha)) \frac{\partial}{\partial A}g(X, v(A_{\tau}); \alpha) - A \right]. 
\end{aligned}
 \label{eq:styleInfer}
 \end{eqnarray} 
Figure \ref{fig:Style_transfer} shows 10 results of style transfer. For each testing image $Y$, we infer $X$ and $C$ by sampling $[X,C] \sim p(X,C|Y)$, which iterates (1) $X \sim p(X|Y,C)$, and (2) $C=v(A)$ where $A \sim p(A|Y,X)$, with randomly initialized $X$ and $C$. We then fix the inferred latent vector $X$, change the category label $C$, and generate images from the combination of $C$ and $X$ by the learned initializer. This experiment demonstrates the effectiveness of our model in style and category disentanglement. 

\subsection{Experiment 2: Image $\rightarrow$  Image}

\subsubsection{Network architecture}
We study learning conditional distributions for image-to-image translation by our framework. The network architectures of the models in this experiment are discussed as follows.

\begin{figure}[h]
\centering	
\includegraphics[width=.85\linewidth]{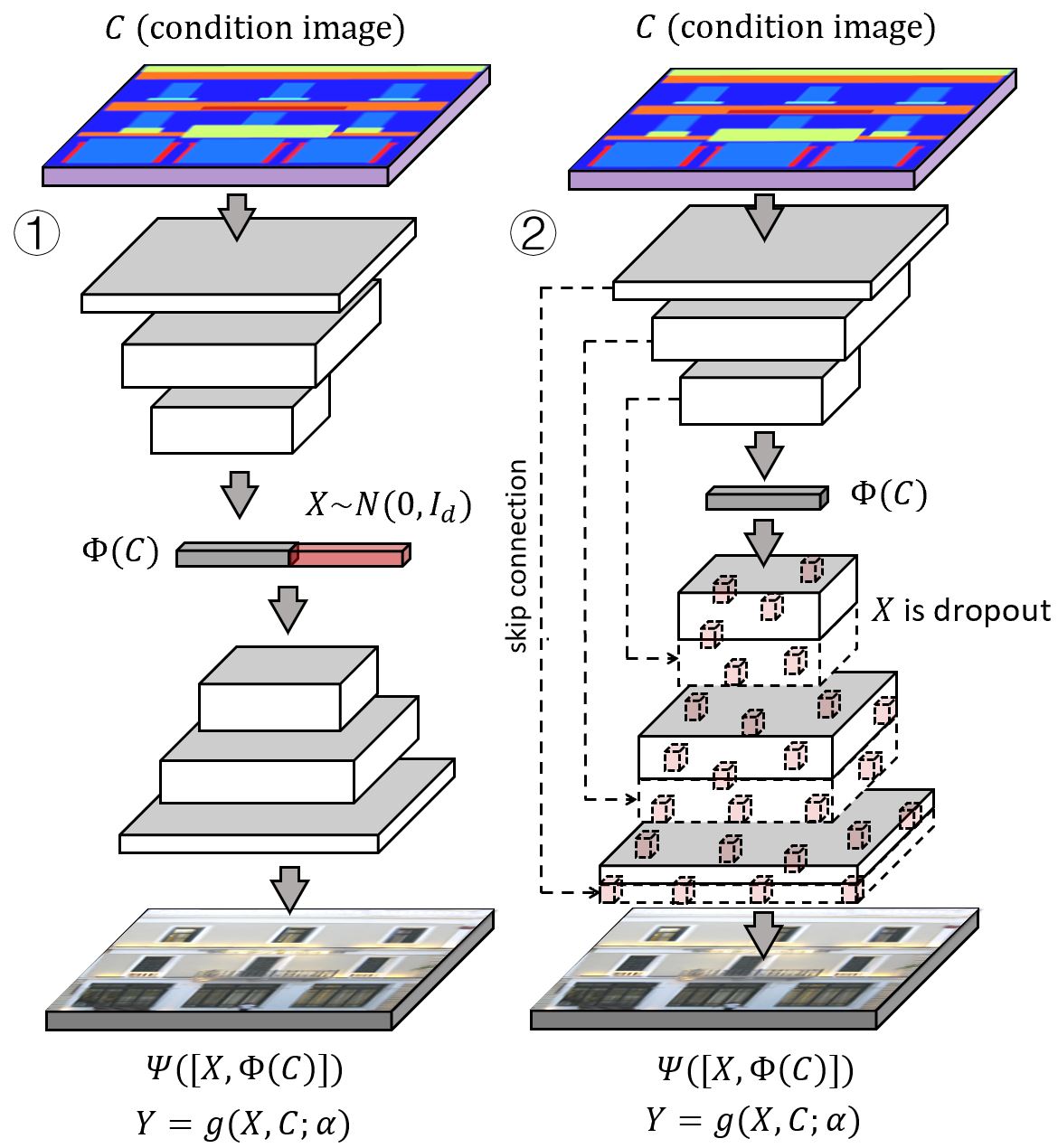}  
\caption{Network architecture of initializer (image-to-image translation). (1) naive straightforward design: the condition image $C$ is first encoded to a vector representation by an encoder $\Phi(C)$, and then the vector is concatenated with the Gaussian noise vector $X$. A decoder $\Psi$ takes as input the concatenated vector $[X, \Phi(C)]$ and outputs the target image $Y$. (2) U-Net with dropout: an encoder-decoder structure ($\Phi$ is the encoder and $\Psi$ is the decoder.), with skip connections added between each layer $j$ and layer $M-j$, where $M$ is the number of layers. Each skip connection concatenates all channels at layer $j$ and those at layer $M-j$. The dropout is applied to each layer in the decoder $\Psi$ to account for randomness $X$.}%
\label{structure_initializer_image}
\end{figure}

As to the initializer, a straightforward design is presented below: we first sample $X$ from the Gaussian noise prior ${\rm N}(0,I_d)$, and we encode the condition image $C$ via an  encoder $\Phi(C)$. The image embedding $\Phi(C)$ is then concatenated to the latent noise vector $X$. After this, we generate target image $Y$ by a decoder $\Psi([X,\Phi(C)])$. The initializer $g(X,C;\alpha)$ is the composition of $\Phi$ and $\Psi$. With Gaussian noise $X$, the initializer will produce stochastic outputs as a distribution. See Figure \ref{structure_initializer_image}(1) for an illustration of the structure. However, in the initial experiments, we found that this design was ineffective in the sense that the generator learned to ignore the noise and produce deterministic outputs. Inspired by \cite{isola2017image}, we design the initializer by following a general shape of the U-Net \cite{ronneberger2015u} with the form of dropout \cite{srivastava2014dropout},  applied on several layers, as noise that accounts for stochasticity in this experiment. A U-Net is an encoder-decoder structure with skip connections added between each layer $j$ and layer $M - j$, where $M$ is the number of layers. Each skip connection performs a concatenation between all channels at layer $j$ and those at layer $M - j$. In the task of image-to-image translation, the input and output images usually differ in appearance but share low-level information. For example, in the case of translating sketch image to photo image, the input and output images are roughly aligned in outline except that they have different colors and textures in appearance. The addition of skip connections allow a direct transfer of low-level information across the network. Figure \ref{structure_initializer_image}(2) illustrates the U-Net structure with dropout as the initializer for image-to-image translation.

\begin{figure}[h]
\centering	
\includegraphics[width=.4\linewidth]{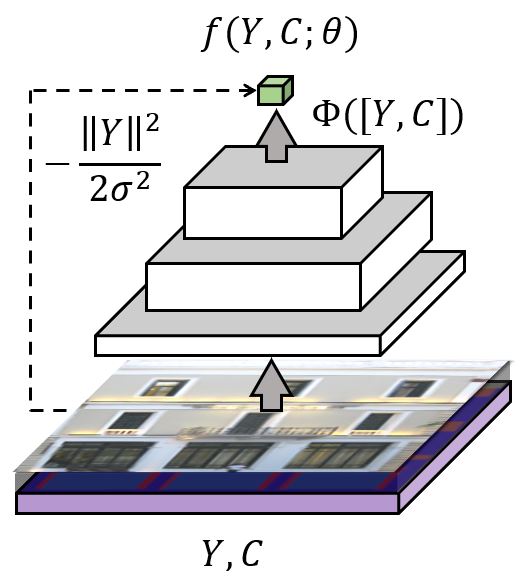}  
\caption{Network architecture of solver (image-to-image translation). Channel concatenation is performed on the condition image $C$ and the target image $Y$. The resulting 6-channel ``image'' is then fed into an encoder $\Phi([Y,C])$. $\Phi$ plus $-\Vert Y\Vert^2/2s^2$ serves as the value function $f(Y,C;\theta)$ in the slow-thinking solver model.}%
\label{structure_solver_image}
\end{figure}

As to the design of the solver model, we first perform channel concatenation on target image $Y$ and condition image $C$, where both images are of the same size. The value function $f(Y,C,\theta)$ is then defined by an encoder $\Phi([Y,C])$ plus $-\Vert Y\Vert^2/2s^2$, in which $\Phi([Y,C])$ maps the 6-channel ``image'' to a scalar by several convolutional layers. Leaky ReLU layers are added between two consecutive convolutional layers. Figure \ref{structure_solver_image} shows an illustration of the network architecture of the solver.

\subsubsection{Semantic labels $\rightarrow$ Scene images }
\label{sec:senmantic}

The experiments are conducted on CMP Facade dataset \cite{tylevcek2013spatial} where each building facade image is associated with an image of architectural labels. The condition image and the target image are of the size of $256 \times 256$ pixels with RGB channels. Data are randomly split into training and testing sets.

In the initializer, the encoder $\Phi$ consists of 8 layers of convolutions with a filter size 4, a subsampling factor 2, and the numbers of channels $\{64, 128, 256, 512, 512, 512, 512, 512\}$ at different layers. Batch normalization and leaky ReLU (with slope 0.2) layers are used after each convolutional layer except that batch normalization is not applied after the first layer. The output of $\Phi$ is 
then fed into $\Psi$, which consists of 8 layers of deconvolutions with a kernel size 4, an up-sampling factor 2, and the numbers of channels $\{512, 512, 512, 512, 256, 128, 64, 3\}$ at different layers. Batch normalization, dropout with a dropout rate of 0.5, and ReLU layer are added between two consecutive deconvolutional layers, and a tanh non-linearity is used after the last layer. The U-Net structure used in this experiment is a connection of the encoder $\Phi$ and the decoder $\Psi$, along with skip connections added to concatenate activations of each layer $j$ and layer $M-j$. ($M$ is the total number of layers.) Therefore, the numbers of output channels of $\Psi$ in the U-Net are $\{1024, 1024, 1024, 1024, 512, 256, 128, 3\}$. The dropout that is applied to each layer of $\Psi$ implies an implicit latent vector $X$ in the initializer. Such an implicit $X$ is too complicated to infer. However, there is no need to infer this $X$ with the cooperative training, which can get around the difficulty of the inference of any complicated forms of latent factors by MCMC teaching. It other words, in each iteration, the learning of the initializer $\Psi([X,\Phi(C)])$ is based on how the MCMC changes the initial examples generated by the initializer from the condition image $C$ and the randomness $X$ due to dropout.

In the solver model, we first perform a channel concatenation on target image $Y$ and condition image $C$, where both images are of size $256 \times 256 \times 3$. 
The value function is then defined by a 4-layer encoder $\Phi([Y,C])$, which maps a 6-channel ``image'' to a scalar as the value score by 3 convolutional layers with
numbers of channels $\{64, 128, 256\}$, filter sizes $\{5, 3, 3\}$ and subsampling
factors $\{2, 2, 1\}$ at different layers (from bottom to top), and one fully connected layer with 100 single filers. Leaky ReLU layer is used between two consecutive convolutional layers. 

Adam is used to optimize the solver with an initial learning rate 0.007, and the initializer with an initial learning rate 0.0001. We set the mini-batch size to be 1. The number of paralleled MCMC chains is also 1. We run 15 Langevin steps with a step size $\delta=0.002$. The standard deviation of the residual in the initializer is $\sigma=0.3$. The standard deviation of the reference distribution in the solver is $s=0.016$. We run 3,000 epochs to train our model.

We adopt random jitter and mirroring for data augmentation in the training stage. As to random jitter, we first resize the input
images from $256 \times 256$  to $286 \times 286$, and then randomly crop image patches with a size $256 \times 256$.   

In this task, we found it beneficial to feed both the refined solutions and the observed ground truth solutions to the initializer, when we update the initializer at each iteration. The solver's job remains unchanged, but the initializer is tasked to not only learn from the solver $\{\tilde{Y}_i\}$ but also to be near the ground truth solutions $\{Y_i\}$. We add an extra $\ell$1 loss to penalize the distance between the output of the initialzer and the ground truth solution. \cite{isola2017image} also finds this strategy effective in training a GAN-based conditional model for image-to-image translation.  

As to the computational time, compared with GAN-based method, our framework has additional $l=15$ steps of Langevin. However, the Langevin is based on gradient, whose computation can be powered by back-propagation, so it is not significantly time-consuming. To be concrete, our method costs
32.7s, while GAN-based method costs 30.9s per epoch for training in a PC with an Intel i7-6700k CPU and a Titan Xp GPU in this experiment.  

\begin{figure*}[h]
\setlength{\tabcolsep}{1pt}
\centering
\begin{tabular}{cccccccc} 
\footnotesize{condition} & \footnotesize{ground truth} & \footnotesize{pix2pix} & \footnotesize{cVAE-GAN} & \footnotesize{cVAE-GAN++} & \footnotesize{BicycleGAN} & \footnotesize{initializer (ours)} & \footnotesize{solver (ours)} \\
\includegraphics[width=.12\linewidth]{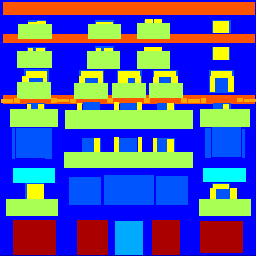}&
\includegraphics[width=.12\linewidth]{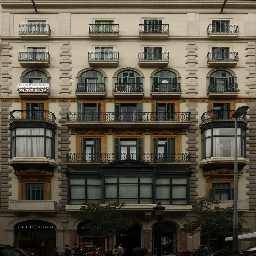}&
\includegraphics[width=.12\linewidth]{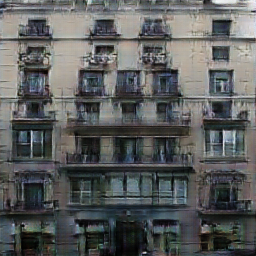}&
\includegraphics[width=.12\linewidth]{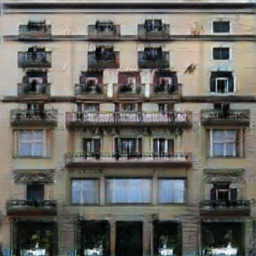}&
\includegraphics[width=.12\linewidth]{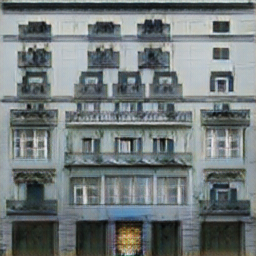}&
\includegraphics[width=.12\linewidth]{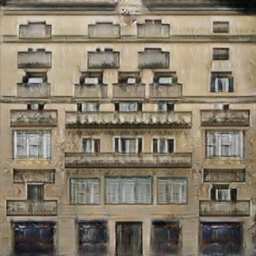}&
\includegraphics[width=.12\linewidth]{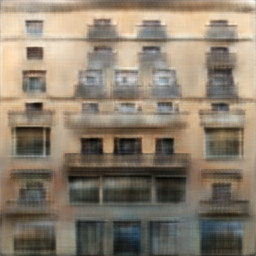}&
\includegraphics[width=.12\linewidth]{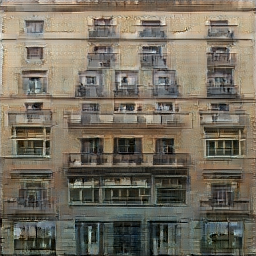}\\ 
\includegraphics[width=.12\linewidth]{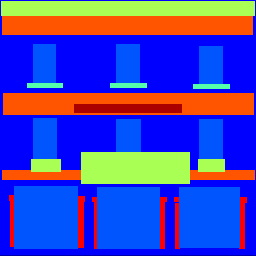}&
\includegraphics[width=.12\linewidth]{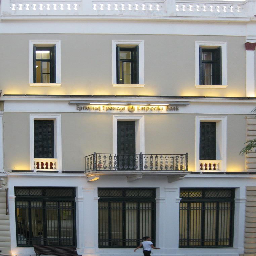}&
\includegraphics[width=.12\linewidth]{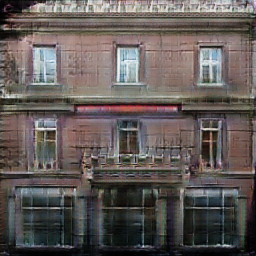}&
\includegraphics[width=.12\linewidth]{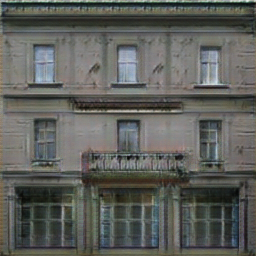}&
\includegraphics[width=.12\linewidth]{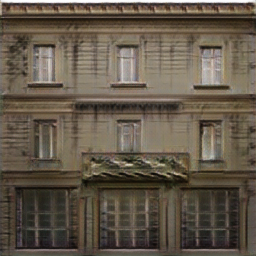}&
\includegraphics[width=.12\linewidth]{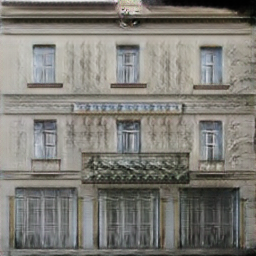}&
\includegraphics[width=.12\linewidth]{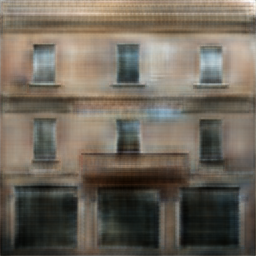}&
\includegraphics[width=.12\linewidth]{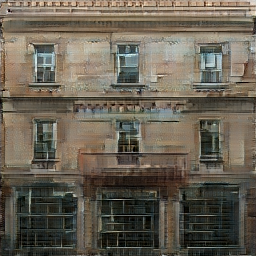}\\ 
\includegraphics[width=.12\linewidth]{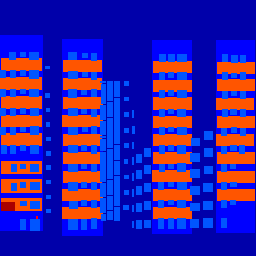}&
\includegraphics[width=.12\linewidth]{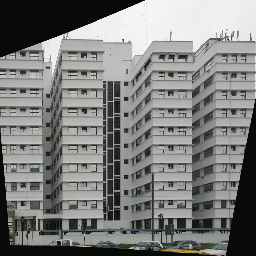}&
\includegraphics[width=.12\linewidth]{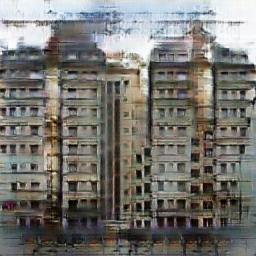}&
\includegraphics[width=.12\linewidth]{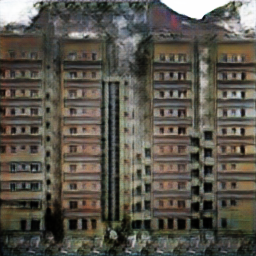}&
\includegraphics[width=.12\linewidth]{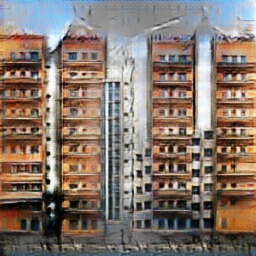}&
\includegraphics[width=.12\linewidth]{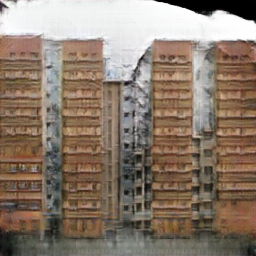}&
\includegraphics[width=.12\linewidth]{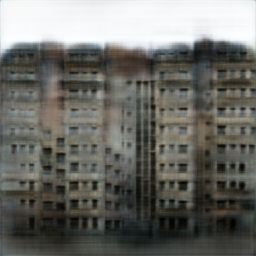}&
\includegraphics[width=.12\linewidth]{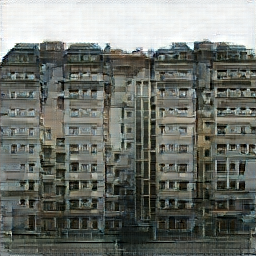}\\ 
\includegraphics[width=.12\linewidth]{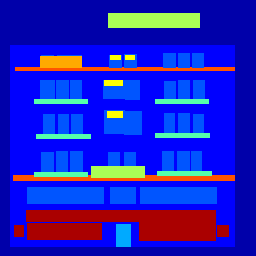}&
\includegraphics[width=.12\linewidth]{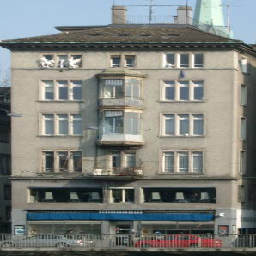}&
\includegraphics[width=.12\linewidth]{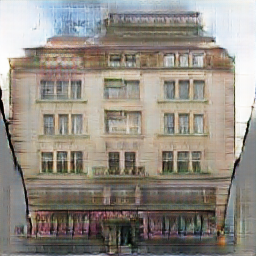}&
\includegraphics[width=.12\linewidth]{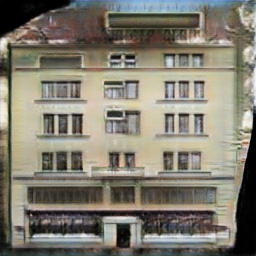}&
\includegraphics[width=.12\linewidth]{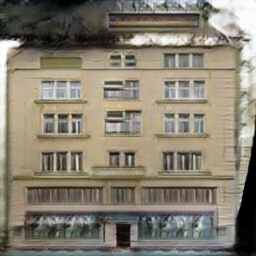}&
\includegraphics[width=.12\linewidth]{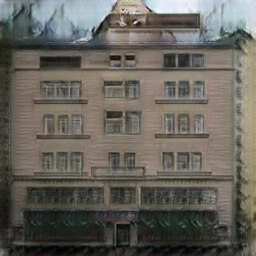}&
\includegraphics[width=.12\linewidth]{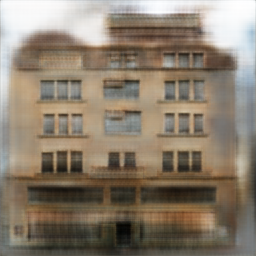}&
\includegraphics[width=.12\linewidth]{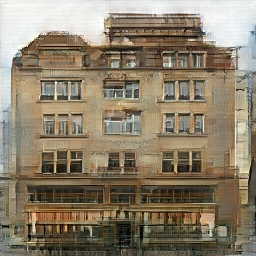}\\ 
\includegraphics[width=.12\linewidth]{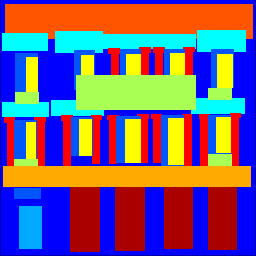}&
\includegraphics[width=.12\linewidth]{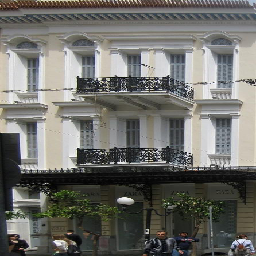}&
\includegraphics[width=.12\linewidth]{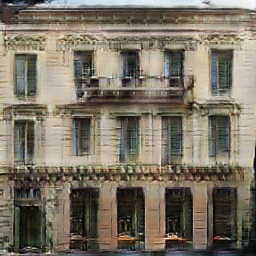}&
\includegraphics[width=.12\linewidth]{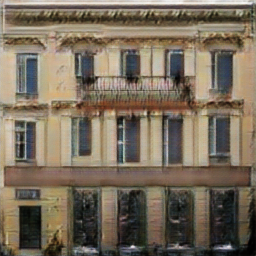}&
\includegraphics[width=.12\linewidth]{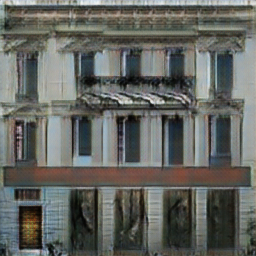}&
\includegraphics[width=.12\linewidth]{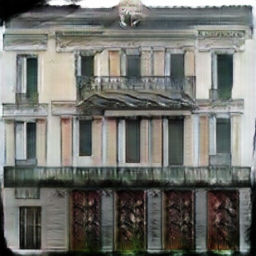}&
\includegraphics[width=.12\linewidth]{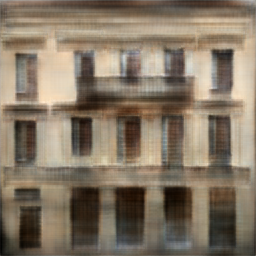}&
\includegraphics[width=.12\linewidth]{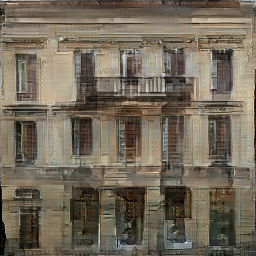}
\end{tabular}

\caption{Generating images conditioned on architectural labels. The first column displays 5 condition images with architectural labels. The second column displays the corresponding ground truth images for reference. For comparison, the third to sixth columns show the generated results by baselines pix2pix, cVAE-GAN, cVAE-GAN++, and BicycleGAN, respectively. The seventh and eighth columns present the generated results obtained by the learned initializer and solver respectively. The training images are of the size $256 \times 256$ pixels. }	
\label{fig:building2}
\end{figure*}

Figure \ref{fig:building2} shows some qualitative results of  generating building facade images from the semantic labels. The first column displays 5 semantic label images that are unseen in the training data. The second column displays the corresponding ground truth images for reference. The results by a baseline method, pix2pix \cite{isola2017image}, are shown in the third row for comparison. pix2pix is a conditional GAN method for image-to-image mapping. Since its generator also uses a U-Net and is paired up with a $\ell$1 loss, for a fair comparison, our initializer adopts exactly the same U-Net structure as in \cite{isola2017image}. The fourth to sixth columns are results generated by some variants of the conditional GAN method, including cVAE-GAN \cite{zhu2017toward}, cVAE-GAN++ \cite{zhu2017toward} and BicycleGAN \cite{zhu2017toward}. The seventh and eighth rows show the generated results conditioned on the semantic label images shown in the first row by the learned initializer and solver, respectively. We can easily observe  qualitative improvements by comparing the outputs of the solver with the ones of the initializer. 

We perform human perceptual tests for evaluating the visual quality of synthesized images. We randomly select 30 different human users to participate in these tests. We compare two methods in each test, where each participant is first presented two images at a time, which are results generated by two different methods given the same conditional input, and then asked which one looks more like a real image. We have total 36 pairwise comparisons in each test for each participant. We evaluate each method by the ratio that the images generated by the method are preferred. As shown in Table \ref{Tab:ratio},  the results generated by our method are considered more realistic by the human subjects. 

\begin{table}[h]
\centering
\begin{small}
\caption{Human perceptual tests for image-to-image synthesis.}%
\label{Tab:ratio}
\begin{tabular}{|l|c|}
\hline
methods & preference ratio \\ \hline \hline
CCoopNets (ours) / cVAE-GAN \cite{zhu2017toward}      & \textbf{0.625} / 0.375       \\ 
CCoopNets (ours) / cVAE-GAN++ \cite{zhu2017toward}& \textbf{0.687} / 0.313 \\ 

CCoopNets (ours) / BicycleGAN \cite{zhu2017toward}& \textbf{0.628} / 0.372      \\
CCoopNets (ours) / pix2pixel \cite{isola2017image} & \textbf{0.720} / 0.280       \\  \hline
\end{tabular}
\end{small}
\end{table}

\subsubsection{Sketch images $\rightarrow$ Photo images}
\label{sec:sketch}
We next test the model on CUHK Face Sketch database (CUFS) \cite{wang2009face}, where for each face, there is a sketch image drawn by an artist based on a photo of the face. We learn to recover the color face images from the sketch images by the proposed framework. The network design and hyperparameter setting are similar to the one we used in Section \ref{sec:senmantic}, except that the mini-batch size and the number of paralleled MCMC chains are set to be 4.

\begin{figure*}[h]
\centering	
\includegraphics[height=.263\linewidth]{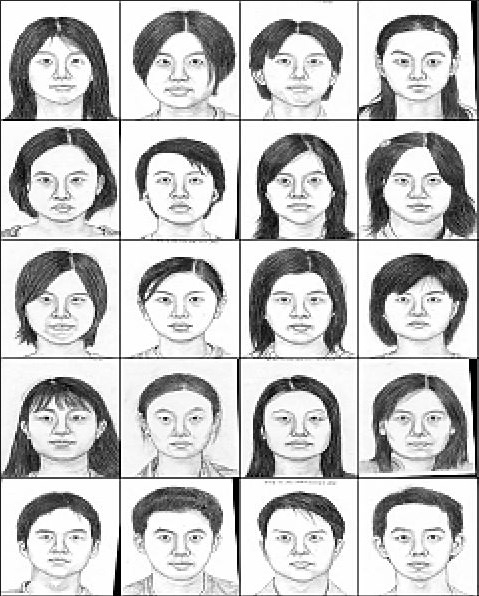}
\includegraphics[height=.263\linewidth]{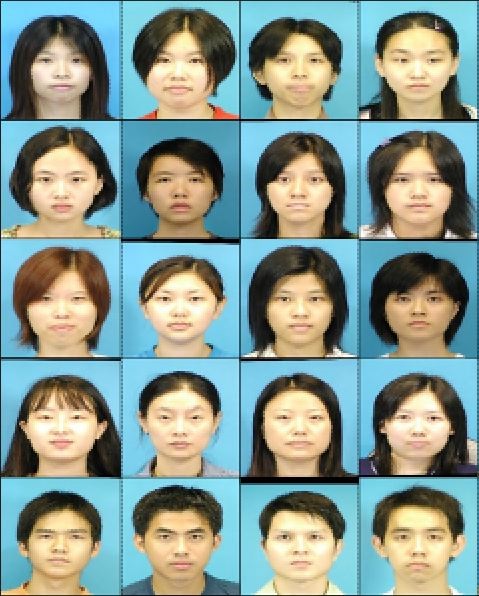}
 \hspace{5mm}
\includegraphics[height=.263\linewidth]{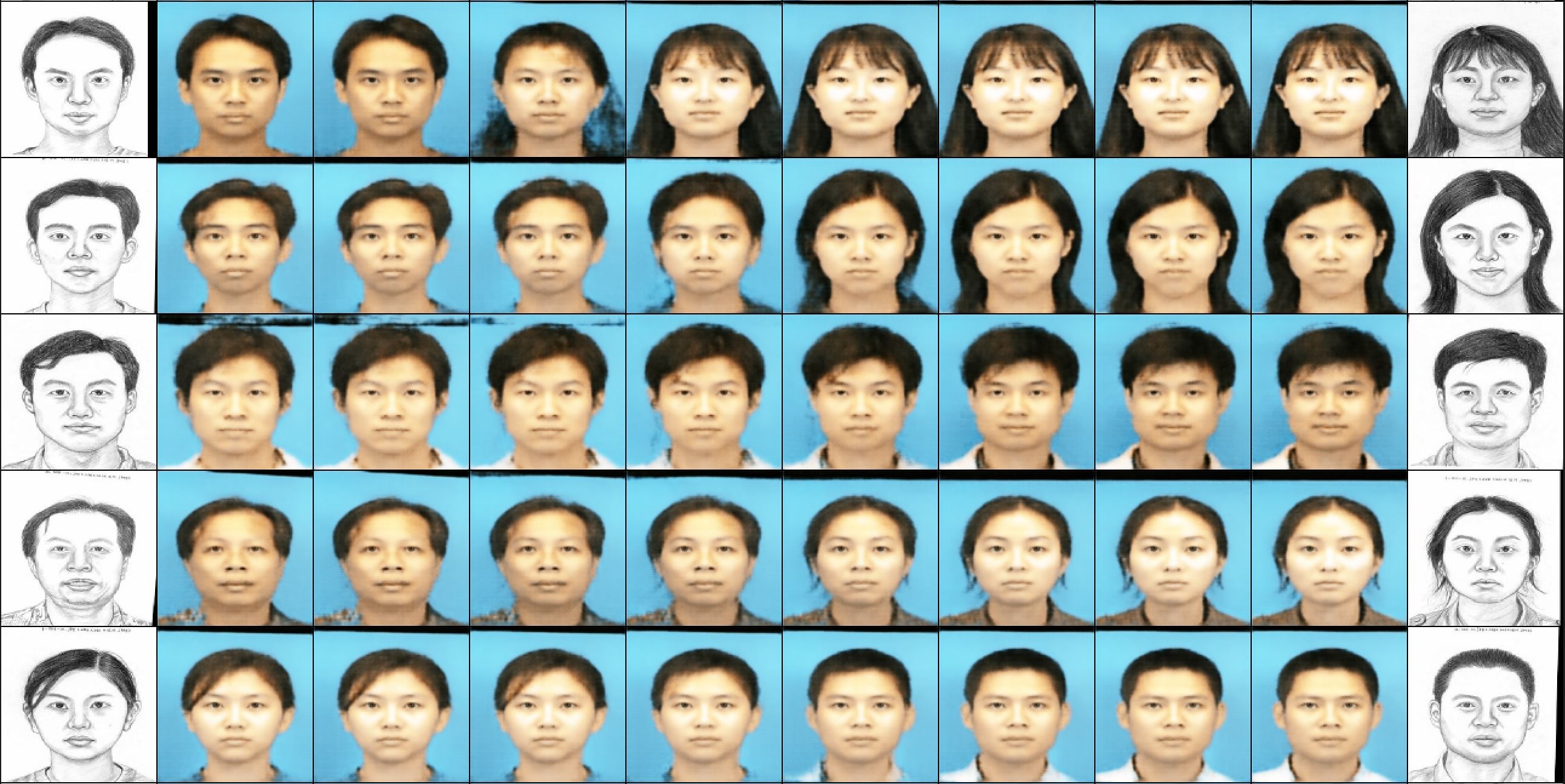} \\
\hspace{-20mm} (a) Recover faces from sketches  \hspace{55mm}(b) Sketch interpolation
\caption{(a) Sketch-to-photo face synthesis. 
Columns 1 through 4: sketch images as conditions. Columns 5 through 8: corresponding face images sampled from the learned models conditioned on sketch images. (b) Sketch interpolation: Generated face images by interpolating between the embedding of the sketch images at two ends, with fixed dropout. Each row displays one example of interpolation.}	
\label{fig:face}
\end{figure*}

Figure \ref{fig:face}(a) displays the face synthesis results conditioned on the sketch images. Columns 1 through 4 show some sketch images as input conditions, while columns 5 through 8 show the corresponding recovered images obtained by sampling from the learned conditional distribution. From the results, we can see that the generated facial appearance (color and texture) in each output image is not only reasonable but also consistent with the input sketch face image in the sense that the face identity in each sketch image remains unchanged after being translating to a photo image.

Figure \ref{fig:face}(b) demonstrates the learned manifold of sketch images (condition) by showing 5 examples of interpolation. For each row, the sketch images at the two ends are first encoded into the embedding by $\Phi(C)$, and then each face image in the middle is obtained by first interpolating the sketch embedding, and then generating the images using the initializer with a fixed dropout, and eventually refining the results by the solver via finite-step Langevin dynamics.  Even though there is no ground truth sketch images for the intervening points, the generated faces appear plausible. Since the dropout $X$ is fixed, the only changing factor is the sketch embedding. We observe smooth changing of the generated faces. 

\begin{figure}[h]
\centering	
\rotatebox{90}{\hspace{0.2mm}{\footnotesize condition}}
\includegraphics[height=.15\linewidth]{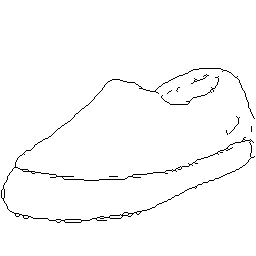}
\includegraphics[height=.15\linewidth]{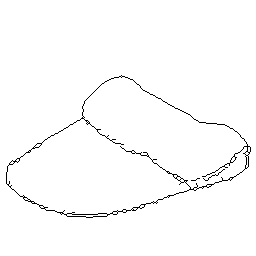}
\includegraphics[height=.15\linewidth]{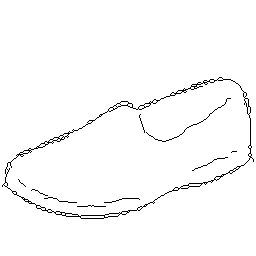}
\includegraphics[height=.15\linewidth]{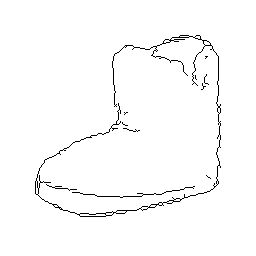}
\includegraphics[height=.15\linewidth]{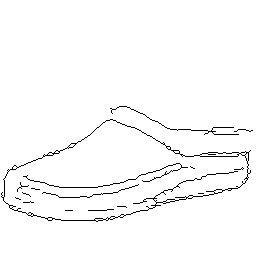}
\includegraphics[height=.15\linewidth]{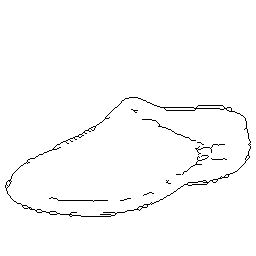}\\

\rotatebox{90}{\hspace{3mm}{\footnotesize GT}}
\includegraphics[height=.15\linewidth]{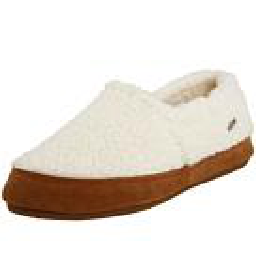}
\includegraphics[height=.15\linewidth]{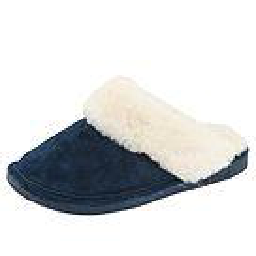}
\includegraphics[height=.15\linewidth]{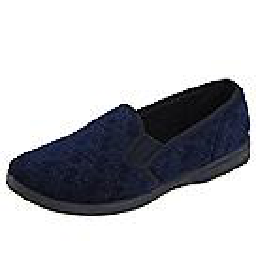}
\includegraphics[height=.15\linewidth]{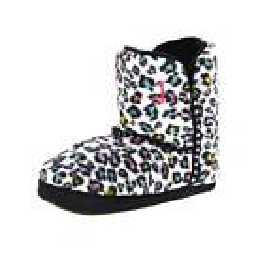}
\includegraphics[height=.15\linewidth]{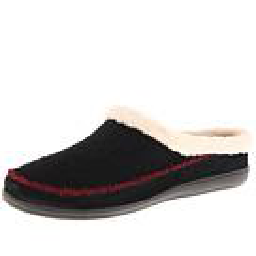}
\includegraphics[height=.15\linewidth]{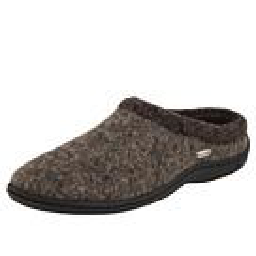}\\

\rotatebox{90}{\hspace{0.5mm}{\footnotesize initializer}}
\includegraphics[height=.15\linewidth]{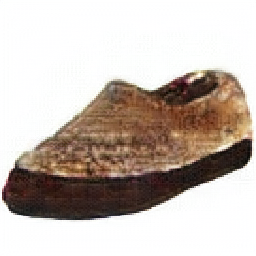}
\includegraphics[height=.15\linewidth]{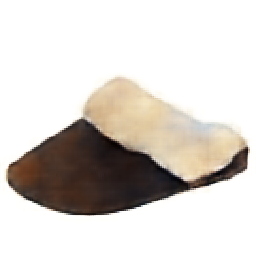}
\includegraphics[height=.15\linewidth]{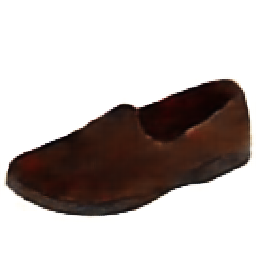}
\includegraphics[height=.15\linewidth]{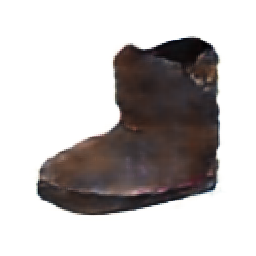}
\includegraphics[height=.15\linewidth]{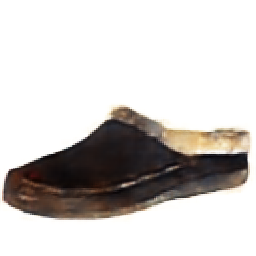}
\includegraphics[height=.15\linewidth]{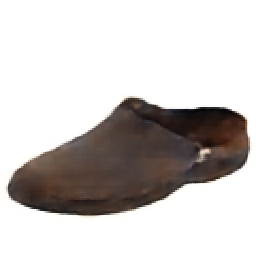}\\
\rotatebox{90}{\hspace{2mm}{\footnotesize solver}}
\includegraphics[height=.15\linewidth]{./results/ZAP50k_test/ZAP50k_test_small/test_000_des.png}
\includegraphics[height=.15\linewidth]{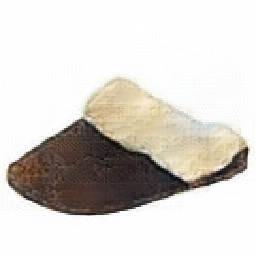}
\includegraphics[height=.15\linewidth]{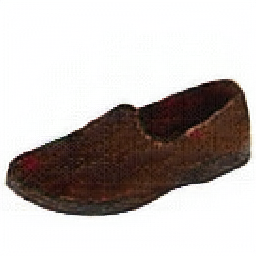}
\includegraphics[height=.15\linewidth]{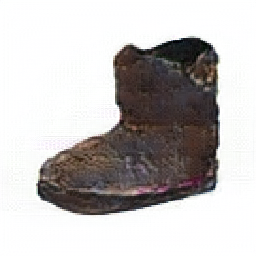}
\includegraphics[height=.15\linewidth]{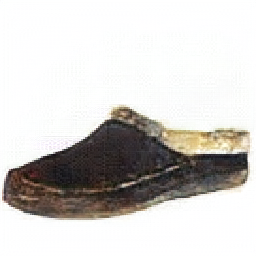}
\includegraphics[height=.15\linewidth]{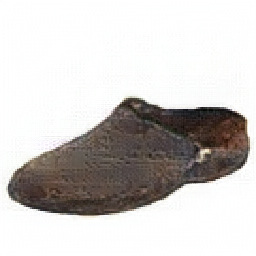}
\caption{Results on edges $\rightarrow$ shoes generation, compared to ground truth. The first row displays the edge images. The second row shows the corresponding ground truth photo images. The last two rows present the generated results obtained by the initializer and the solver, respectively.}	
\label{fig:shoe}
\end{figure}

We conduct another experiment on UT Zappos50K dataset \cite{tylevcek2013spatial} for photo image recovery from edge image. The dataset contains $50k$ training images of shoes. Edge images are computed by HED edge detector \cite{xie2015holistically} with post processing. We use the same model structure as the one in the last experiment. Figure \ref{fig:shoe} shows some qualitative results of synthesizing shoe images from edge images.

\subsubsection{Image inpainting}

We also test our method on the task of image inpainting by learning a mapping from an occluded image (256 $\times$ 256 pixels), where a mask with the size of $128 \times 128$ pixels is centrally placed onto the original version, to the original image. We use Paris streetview \cite{pathak2016context} and the CMP Facade dataset. In this case, $C$ is the observed part of the input image, and $Y$ is the unobserved part of the image. The network architectures for both initializer and solver, along with hyperparameter setting, are similar to those we used in Section \ref{sec:senmantic}. To recover the occluded part of the input images, we only update the pixels of the occluded region in the Langevin dynamics. 

We compare our method with some baselines, including pix2pix, cVAE-GAN, cVAE-GAN++ and BicycleGAN. Table~\ref{tab:inpainting} shows quantitative results where the recovery performance is measured by the peak signal-to-noise ratio (PSNR) and structural similarity measures (SSIM), which are computed between the occlusion regions of the generated example and the ground truth example. The batch size is one. Our method outperforms the baseline methods using adversarial training or variational inference in this recovery task. Table \ref{tab:inpainting_para} reports a comparison of model complexity with the baseline methods on CMP Facade dataset in terms of number of model parameters and running time.

\begin{figure*}[h]
\setlength{\tabcolsep}{1pt}
\centering	
\begin{tabular}{cccccccc}
\footnotesize{condition} & \footnotesize{ground truth} & \footnotesize{pix2pix} & \footnotesize{cVAE-GAN} & \footnotesize{cVAE-GAN++} & \footnotesize{BicycleGAN} & \footnotesize{initializer (ours)} & \footnotesize{solver (ours)} \\
\includegraphics[width=.12\linewidth]{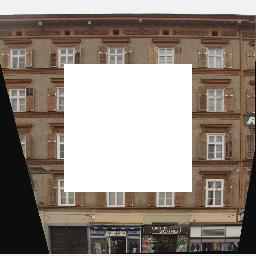}&
\includegraphics[width=.12\linewidth]{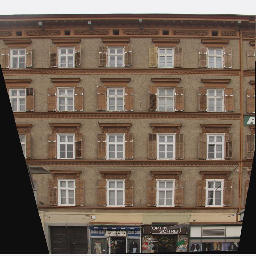}&
\includegraphics[width=.12\linewidth]{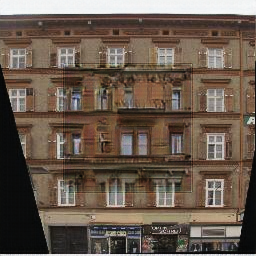}&
\includegraphics[width=.12\linewidth]{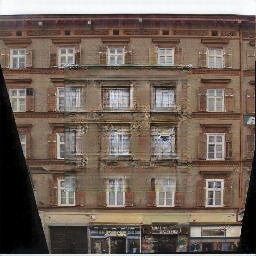}&
\includegraphics[width=.12\linewidth]{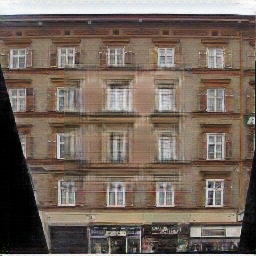}&
\includegraphics[width=.12\linewidth]{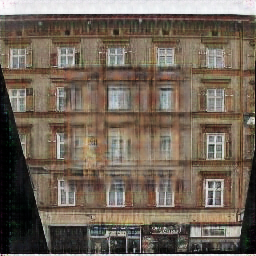}&
\includegraphics[width=.12\linewidth]{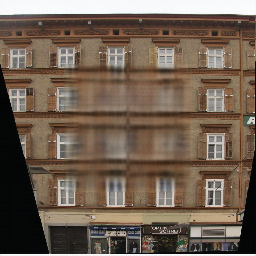}&
\includegraphics[width=.12\linewidth]{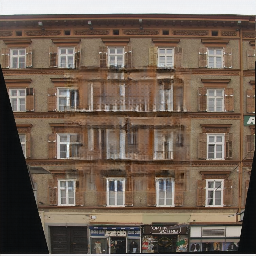}
\\

\includegraphics[width=.12\linewidth]{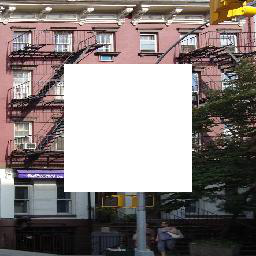}&
\includegraphics[width=.12\linewidth]{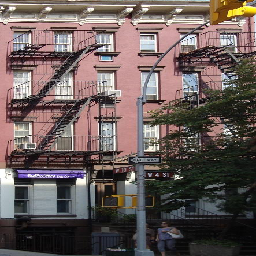}&
\includegraphics[width=.12\linewidth]{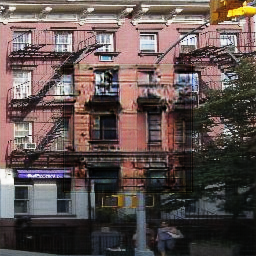}&
\includegraphics[width=.12\linewidth]{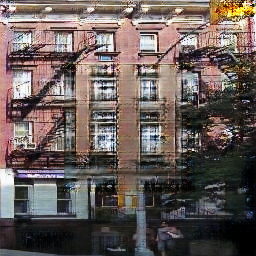}&
\includegraphics[width=.12\linewidth]{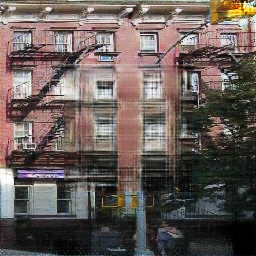}&
\includegraphics[width=.12\linewidth]{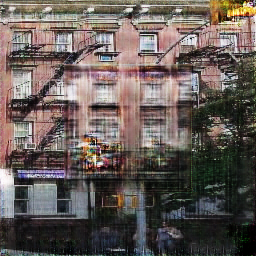}&
\includegraphics[width=.12\linewidth]{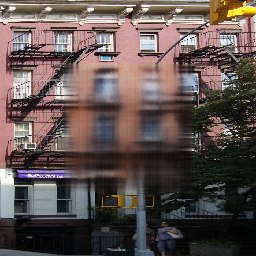}&
\includegraphics[width=.12\linewidth]{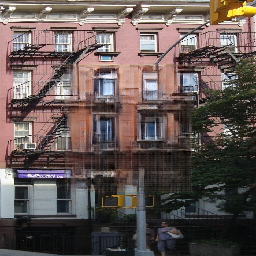}
\\
\includegraphics[width=.12\linewidth]{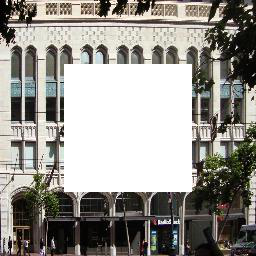}&
\includegraphics[width=.12\linewidth]{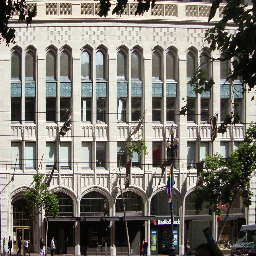}&
\includegraphics[width=.12\linewidth]{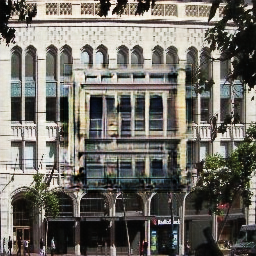}&
\includegraphics[width=.12\linewidth]{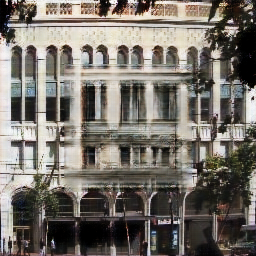}&
\includegraphics[width=.12\linewidth]{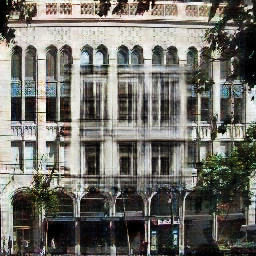}&
\includegraphics[width=.12\linewidth]{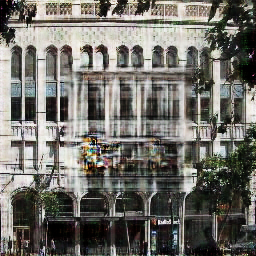}&
\includegraphics[width=.12\linewidth]{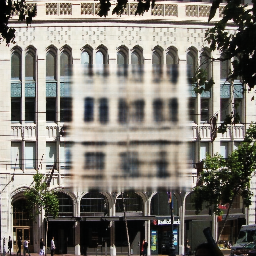}&
\includegraphics[width=.12\linewidth]{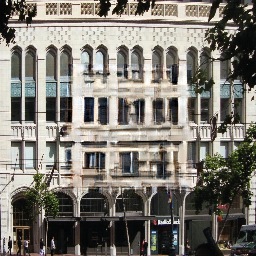}
\\
\includegraphics[width=.12\linewidth]{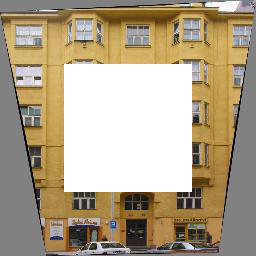}&
\includegraphics[width=.12\linewidth]{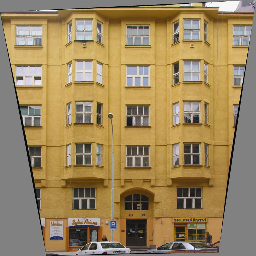}&
\includegraphics[width=.12\linewidth]{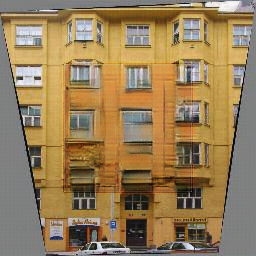}&
\includegraphics[width=.12\linewidth]{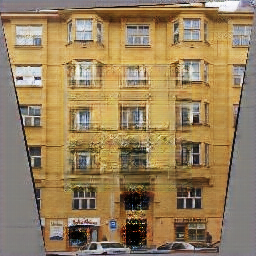}&
\includegraphics[width=.12\linewidth]{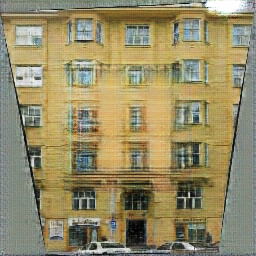}&
\includegraphics[width=.12\linewidth]{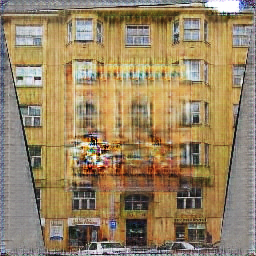}&
\includegraphics[width=.12\linewidth]{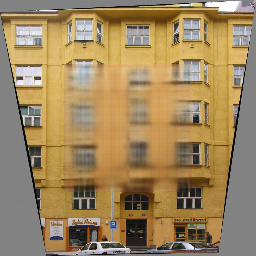}&
\includegraphics[width=.12\linewidth]{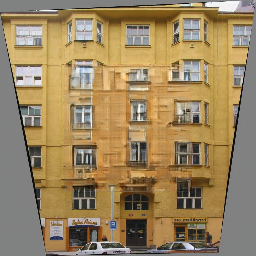}\\
\end{tabular}
\caption{Results of photo inpainting. Each row displays one example. The first image is the testing image ($256 \times 256$ pixels) with a hole of $128 \times 128$ pixels that needs to be recovered, the second image is the ground truth, the third to sixth images are the results recovered by pix2pix, cVAE-GAN, cVAE-GAN++, BicycleGAN for comparison. The seventh and the last images are the results recovered by the initializer and the solver, respectively.}
\label{fig:recovery}
\end{figure*}

Figure \ref{fig:recovery} shows a comparison of qualitative results of different methods on CMP Facade dataset. Each row displays one example. The first image is the testing image with a hole that needs to be recovered. The second image is the ground truth image. The third to sixth images are the inpainting results obtained by pix2pix, cVAE-GAN, cVAE-GAN++ and BicycleGAN, respectively. The seventh and the last images are the results recovered by the initializer and the solver, respectively.

\begin{table}[h]
\centering
\begin{small}
\caption{Comparison with the baseline methods for image inpainting on the CMP Facade dataset and Paris streetview dataset.}%
\label{tab:inpainting}
\begin{tabular}{|l|c|c|c|c|}
\hline
\multirow{2}{*}{Model} & \multicolumn{2}{c|}{CMP Facades}                      & \multicolumn{2}{c|}{Paris streetview}                                \\ \cline{2-5} 
                  & \multicolumn{1}{l|}{PSNR} & \multicolumn{1}{l|}{SSIM} & \multicolumn{1}{l|}{PSNR} & \multicolumn{1}{l|}{SSIM} \\ \hline \hline
cVAE-GAN \cite{zhu2017toward} & 19.43 & 0.68 & 16.12 & 0.72 \\
cVAE-GAN++ \cite{zhu2017toward} & 19.14 & 0.64 &16.03 &0.69 \\
BicycleGAN \cite{zhu2017toward} & 19.07 & 0.64 & 16.00 & 0.68 \\
pix2pix\cite{isola2017image} & 19.34 & 0.74 & 15.17  & 0.75 \\ 
CCoopNets (ours)  & \textbf{20.47} & \textbf{0.77} & \textbf{21.17} &\textbf{0.79} \\ \hline
\end{tabular}
\end{small}
\end{table}

\begin{table}[h]
\centering
\begin{small}
\caption{Comparison of model complexity with the baseline methods for image inpainting on CMP Facade dataset.}%
\label{tab:inpainting_para}
\begin{tabular}{|l|c|c|}
\hline
\multirow{2}{*}{Model} & Size & Time \\
\cline{2-3} 
& $\sharp$ of parameters & sec / epoch  \\ \hline \hline
cVAE-GAN \cite{zhu2017toward} & 60.85M &  12.06 \\
cVAE-GAN++ \cite{zhu2017toward} & 64.30M & 18.40  \\
BicycleGAN \cite{zhu2017toward} & 64.30M &  25.60 \\
pix2pix \cite{isola2017image} & 57.89M &  12.62 \\ 
CCoopNets (ours)  & 55.84M & 22.43 \\ \hline
\end{tabular}
\end{small}
\end{table}

\section{Conclusion}
Solving a challenging problem usually requires an iterative algorithm. This amounts to slow thinking. The iterative algorithm usually requires a good initialization to jumpstart it so that it can converge quickly. The initialization amounts to fast thinking. For instance, reasoning and planning usually require iterative search or optimization, which can be initialized by a learned computation in the form of a neural network. Thus integrating fast thinking initialization and slow thinking sampling or optimization is very compelling. This paper addresses the problem of high-dimensional conditional learning and proposes a cooperative learning method that couples a fast thinking initializer and a slow thinking solver. The initializer initializes the iterative optimization or sampling process of the solver, while the solver in return teaches the initializer by distilling its iterative algorithm into the initializer. We demonstrate the proposed method on a variety of image synthesis and recovery tasks. Compared to GAN-based method, such as conditional GANs, our method is equipped with an extra iterative sampling and optimization algorithm to refine the solution, guided by a learned objective function. This may prove to be a powerful method for solving challenging conditional learning problems.



\ifCLASSOPTIONcompsoc
  \section*{Acknowledgments}
\else
  \section*{Acknowledgment}
\fi

We gratefully acknowledge the support of NVIDIA Corporation with the donation of the Titan Xp GPU used for this research. The work is supported by NSF DMS-2015577, DARPA SIMPLEX N66001-15-C-4035, ONR MURI N00014-16-1-2007, DARPA ARO W911NF-16-1-0579, DARPA N66001-17-2-4029, and XSEDE grant ASC180018.
\ifCLASSOPTIONcaptionsoff
  \newpage
\fi



%

\bibliographystyle{IEEEtran}
\bibliography{mybibfile}

%
%

%




\end{document}